\documentclass[journal,twoside,web]{ieeecolor}
\usepackage{tmi}
\usepackage{cite}
\usepackage{amsmath,amssymb,amsfonts}
\usepackage{algorithmic}
\usepackage{graphicx}
\usepackage{textcomp}
\usepackage{multirow}
\usepackage{colortbl}
\usepackage[table]{xcolor}
\usepackage{xcolor}
\usepackage{arydshln}
\usepackage{url}
\usepackage{bbding}
\usepackage{makecell}
\usepackage{subcaption}
\usepackage{caption} 
\usepackage{hyperref}
\usepackage{booktabs}
\usepackage{multirow}
\usepackage{makecell}
\usepackage[table]{xcolor}
\usepackage{adjustbox}

\newcommand{\wci}[3]{\makecell[c]{#1\\[-2.5pt]{\scriptsize\color{gray}[#2,\,#3]}}}
\usepackage[font=footnotesize,labelfont=bf]{caption}
\def\BibTeX{{\rm B\kern-.05em{\sc i\kern-.025em b}\kern-.08em
    T\kern-.1667em\lower.7ex\hbox{E}\kern-.125emX}}
\begin{document}
\title{Bridging the Gap in Ophthalmic AI: MM-Retinal-Reason Dataset and OphthaReason Model toward Dynamic Multimodal Reasoning}
\author{Ruiqi Wu, Yu’ang Yao, Na Su, Tengfei Ma, Chenran Zhang, Tao Zhou, \IEEEmembership{Senior Member, IEEE},\\ Tianyu Mao, Ye Geng, Nianfeng Tang, Xiangyuan Duanmu, Qiqi Chen,\\ Geng Chen, \IEEEmembership{Senior Member, IEEE}, Wen Fan$^*$, Yi Zhou$^*$, \IEEEmembership{Senior Member, IEEE}
\thanks{This work was supported by the National Natural Science Foundation of China under Grant 62476054, Grant 82401284, and Grant 62576153.}
\thanks{R. Wu, Y. Yao, T. Ma, C. Zhang, Q. Chen, and Y. Zhou are with the School of Computer Science and Engineering, Southeast University, Nanjing, China.}
\thanks{N. Su, T. Mao, Y. Geng, N. Tang, X. Duanmu, and W. Fan are with the Department of Ophthalmology, The First Affiliated Hospital of Nanjing Medical University, Nanjing, China.}
\thanks{T. Zhou is with the School of Computer Science and Engineering, Nanjing University of Science and Technology, Nanjing, China.}
\thanks{G. Chen is with the School of Computer Science, Northwestern Polytechnical University, Xi’an, China.}}

\maketitle
\def\thefootnote{*}\footnotetext{Corresponding authors: \textit{Wen Fan} (fanwen@njmu.edu.cn) and \textit{Yi Zhou} (yizhou.szcn@gmail.com).}\def\thefootnote{\arabic{footnote}}

\begin{abstract}
Multimodal large language models (MLLMs) have recently demonstrated remarkable reasoning abilities under reinforcement learning (RL) paradigm. However, most existing multimodal medical reasoning models focus on basic reasoning, which refers to shallow inference based on visual feature matching. In contrast, real-world clinical diagnosis extends beyond basic reasoning, demanding complex reasoning that integrates heterogeneous clinical information (such as chief complaints and medical history) with multimodal medical imaging data. To bridge this gap, we introduce MM-Retinal-Reason, {an ophthalmic multimodal dataset covering the full spectrum of perception and reasoning. Specifically, it is the first dataset in ophthalmology to encompass both basic and complex reasoning tasks with Chain-of-Thought (CoT) trajectories,} aiming to enhance visual-centric reasoning and emulate realistic clinical decision-making. Building upon MM-Retinal-Reason, we propose OphthaReason, the first {RL-enhanced} ophthalmic multimodal reasoning model with step-by-step reasoning traces. To enable flexible adaptation to both basic and complex reasoning tasks, we further introduce Uncertainty-Aware Dynamic Thinking (UADT), which estimates sample-level uncertainty via entropy and dynamically modulates exploration depth through a shaped advantage mechanism. Comprehensive experiments demonstrate the effectiveness of our model on both basic and complex reasoning tasks, outperforming general-purpose MLLMs, medical MLLMs, RL-based medical MLLMs, and ophthalmic MLLMs by at least 15.47\%. Project Page: \href{https://github.com/lxirich/OphthaReason}{link}.
\end{abstract}

\begin{IEEEkeywords}
Multimodal Reasoning, Reinforcement Learning, Ophthalmology, Dynamic Reasoning
\end{IEEEkeywords}

\section{Introduction}

\IEEEPARstart{M}{edical} diagnosis is a complex process that fundamentally relies on logical inference and clinical expertise. In ophthalmology, this challenge is more pronounced, demanding the integration of diverse imaging modalities and extensive clinical contexts. Recent advancements in multimodal large language models (MLLMs) 
\cite{jaech2024openai}\cite{guo2025deepseek}\cite{deng2025openvlthinker}\cite{tan2025reason}\cite{liu2025visual} offer a powerful solution. By incorporating explicit reasoning traces, reinforcement learning (RL)-based models can significantly improve interpretability and reliability thus fostering safer and more trustworthy medical artificial intelligence (AI).

\begin{figure}[!t]
\centerline{\includegraphics[width=\columnwidth]{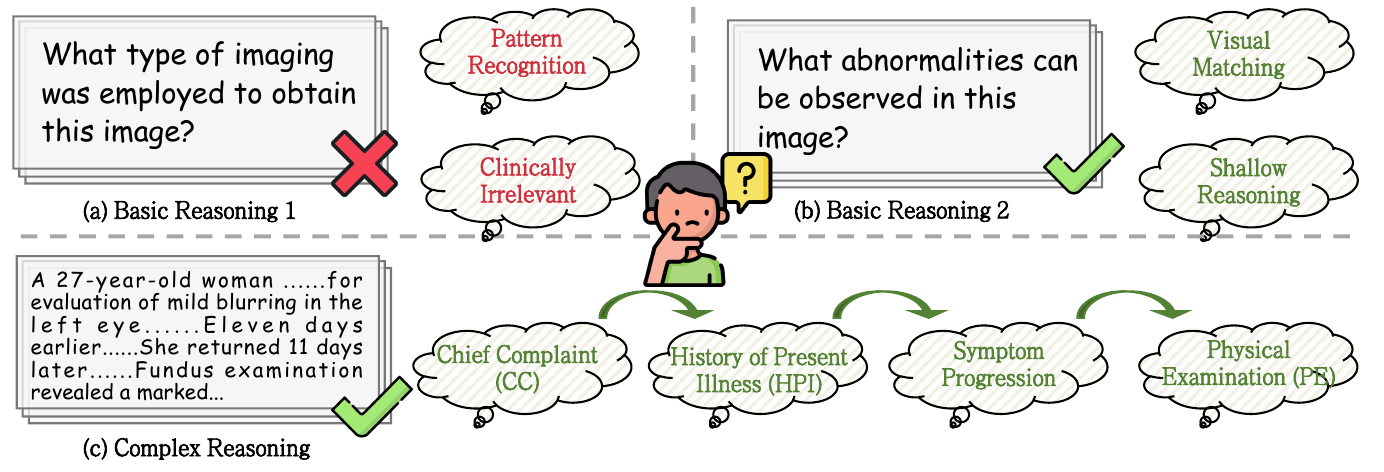}}
\caption{Comparison between basic and complex reasoning tasks. (a) Some basic tasks focus on modality or organ recognition, which are clinically irrelevant. (b) Other basic tasks pertain to diagnosis, involving shallow reasoning due to limited information in the question. (c) Complex tasks incorporate heterogeneous real-world clinical information, enabling deeper reasoning. Our dataset includes both (b) and (c) tasks, and OphthaReason dynamically adapts to varying reasoning complexities.}
\label{fig:diff_in_dataset}
\end{figure}

Most existing multimodal medical reasoning models \cite{pan2025medvlm}\cite{lai2025med}\cite{fan2025chestx}\cite{ zhang2025patho} are tailored for general medicine, pathology and radiology. However, due to the lack of large-scale and well-annotated ophthalmology-specific visual question answering (VQA) data, the development of ophthalmic reasoning models falls behind. Moreover, existing models primarily focus on basic tasks with simple queries as illustrated in Fig.~\ref {fig:diff_in_dataset}(a)(b). The former often involves questions of limited clinical relevance, such as imaging modality recognition and orientation identification, while the latter relies on visual feature matching with shallow reasoning. Though effective for basic diagnoses, such models often fail to address the complex reasoning required in real clinical practice. Unlike standardized benchmarks, real-world clinical scenarios involve ambiguous symptoms and comorbidities that require more than knowledge matching on a single imaging modality \cite{thapa2025disentangling}. Accurate diagnosis often requires diverse imaging modalities and heterogeneous clinical information, such as chief complaint (CC), history of present illness (HPI), symptom progression, and physical examination (PE) \cite{pfob2022importance} (see Fig.~\ref {fig:diff_in_dataset} (c)), which is not only helpful in diagnosing common diseases but essential when confronting complex or rare conditions.

These limitations underscore the necessity of developing multimodal medical reasoning models that can handle multi-source, heterogeneous reasoning tasks with varying levels of diagnostic complexity. However, most existing approaches treat all reasoning tasks indiscriminately \cite{rui2025improving}\cite{fan2025chestx}\cite{lai2025med}\cite{zhang2025patho}, without adapting the model’s exploratory capabilities to varying task complexity. This limits their ability to cope with heterogeneous diagnostic scenarios, resulting in low reasoning efficiency and inadequate exploration depth. Although QoQ-Med \cite{dai2025qoq} attempts to address the challenge of heterogeneous data by framing it as a domain imbalance problem, its effectiveness is constrained by reliance on pre-defined domains and a coarse-grained, cluster-level estimation strategy, underscoring the need for reasoning models with fine-grained, sample-level dynamic thinking capabilities.

To address these challenges, we introduce MM-Retinal-Reason, {an ophthalmology-specific multimodal reasoning dataset} designed for expert-level diagnostic tasks, integrating both basic and complex reasoning tasks to reflect real-world clinical scenarios. MM-Retinal-Reason is built from real-world data collected from 44 public datasets and PubMed Central (PMC), supplemented with reasoning trajectories. It comprises 4 types of questions, including true/false questions, single-answer multiple-choice, multiple-answer multiple-choice, and open-ended questions, covering over 100 common and rare ophthalmic conditions. MM-Retinal-Reason consists of four components: basic reasoning VQA, chain-of-thought (CoT) reasoning trajectories, image-caption pairs, and complex reasoning VQA, providing comprehensive supervision for the training and evaluation of ophthalmic MLLMs across the full spectrum of perception and reasoning.

Furthermore, we propose OphthaReason, a multimodal medical reasoning model for ophthalmic diagnosis that supports dynamic reasoning across tasks ranging from basic questions to complex scenarios. It adopts a three-stage training pipeline consisting of vision-language alignment, CoT supervised fine-tuning (SFT), and reinforcement learning, which together enable perception enhancement, stepwise reasoning, and clinical decision-making. Specifically, we develop Uncertainty-Aware Dynamic Thinking (UADT), which leverages sample-wise entropy to guide adaptive reasoning. Based on each sample’s uncertainty, UADT encourages broader exploration of uncertain challenging cases while mitigating excessive updates on confident trivial ones. This mechanism allows OphthaReason to simulate the flexible cognitive process of human experts, dynamically adjusting exploration depth to match the complexity of each sample.

In summary, our key contributions are highlighted as: (1) The construction of MM-Retinal-Reason, which is the first multimodal ophthalmic dataset designed for both basic and complex reasoning tasks, provides comprehensive supervision across perceptual understanding and cognitive reasoning. (2) We introduce OphthaReason, which is the first RL-enhanced ophthalmic multimodal reasoning model to the best of our knowledge. It is trained through a three-stage pipeline to enhance both perceptual understanding and reasoning capability. (3) The creation of Uncertainty-Aware Dynamic Thinking dynamically promotes deeper exploration for uncertain samples while applying minimal updates to certain ones based on sample-level entropy. (4) Extensive experiments demonstrate that OphthaReason consistently enhances performance across diverse imaging modalities and varying reasoning complexity, achieving at least 15.47\% improvements over general-purpose, medical, RL-based medical, and ophthalmic MLLMs.

\section{Related Work}
\subsection{Multimodal Medical and Ophthalmic Datasets}

Early medical VQA benchmarks, including VQA-Med \cite{ben2019vqa}, PMC-VQA \cite{zhang2023pmc}, and OmniMedVQA \cite{hu2024omnimedvqa}, mainly target basic single-image questions like disease or modality identification, failing to capture the contextual complexity of real-world clinical diagnosis. Recent benchmarks have broadened this scope. MedXpertQA-MM \cite{zuo2025medxpertqa} provides expert-level questions but lacks explicit reasoning traces, while MedFrameQA \cite{yu2025medframeqa} supports multi-image reasoning but omits key clinical context such as patient history and examination findings. {Ophthalmic datasets have also advanced multimodal learning, but important limitations remain. LMOD \cite{qin2025lmod} is built mainly from structured annotations rather than native generative VQA. Eyecare-100K \cite{li2025eyecaregpt} supports multimodal QA and report generation but lacks explicit CoT and context-rich multi-image reasoning. FundusGen \cite{liu2025constructing} provides CFP-only, multi-turn cognitive chains rather than explicit CoT or complex reasoning across multiple imaging modalities.} To fill this gap, we propose MM-Retinal-Reason, spanning from basic to complex reasoning tasks with explicit reasoning trajectories.

\begin{figure*}[!t]
\centerline{\includegraphics[scale=0.55]{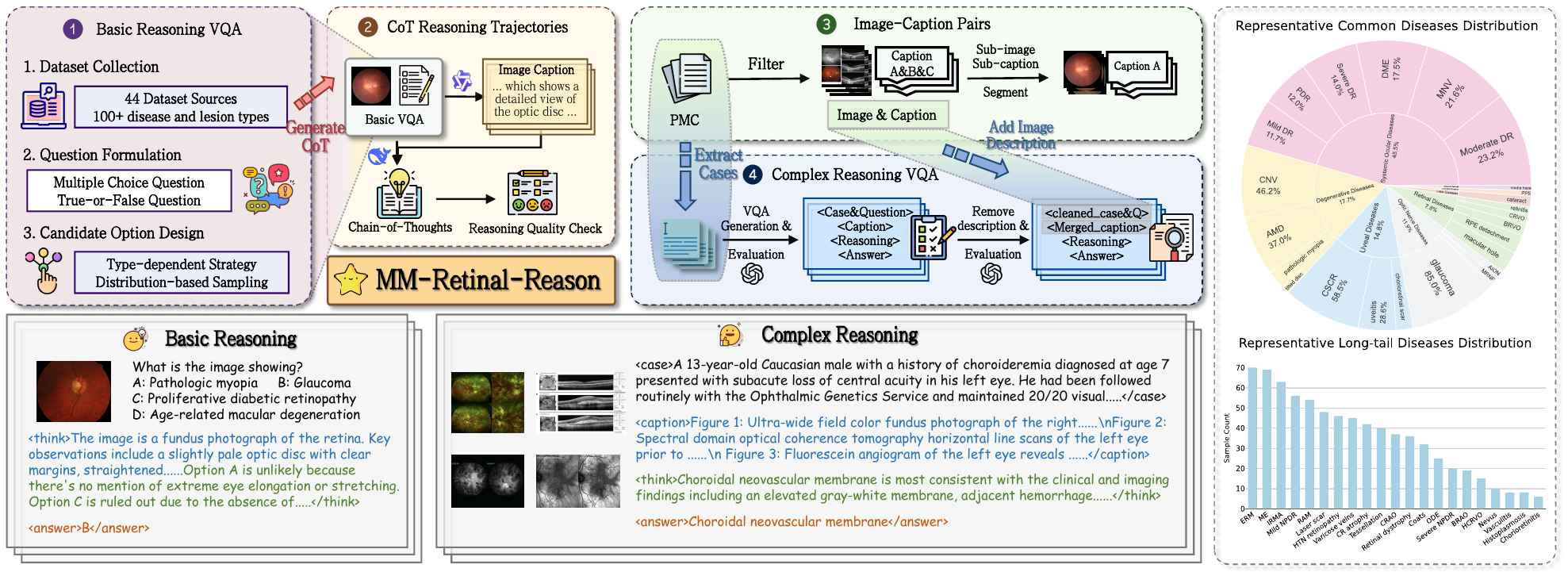}}
\caption{{\textbf{Curation workflow and comprehensive diversity of MM-Retinal-Reason.} MM-Retinal-Reason consists of four interconnected components covering a wide range of perception and reasoning tasks. Example basic and complex VQA cases are shown at the bottom. The category distribution is illustrated on the right, where the upper chart highlights representative common disease categories with larger sample sizes and the lower chart presents representative long-tail disease categories with limited samples. Disease names are abbreviated for readability. This design simulates the full spectrum of real-world clinical challenges while ensuring robust model generalization.}}
\label{fig:dataset_category}
\end{figure*}

\subsection{Multimodal Medical Reasoning Models with RL}
Reinforcement learning (RL) has played a pivotal role in advancing multimodal medical reasoning. These models can be broadly categorized into two streams: generalist models like MedVLM-R1 \cite{pan2025medvlm}, Med-R1\cite{lai2025med}, QoQ-Med\cite{dai2025qoq}, lingshu-RL \cite{xu2025lingshu}, MedCCO \cite{rui2025improving}, which target broad medical reasoning tasks, and specialist models tailored to specific clinical domains such as ChestX-Reasoner \cite{fan2025chestx}, BoxMed-RL \cite{jing2025reason} for radiology and Patho-R1 \cite{zhang2025patho}, PathVLM-R1 \cite{wu2025pathvlm} for Pathology. Nonetheless, these advancements have not yet been extended to ophthalmology. To bridge this gap, we introduce OphthaReason, an RL-enhanced ophthalmic reasoning model. Unlike prior approaches that apply uniform exploration strategies across all samples, OphthaReason dynamically modulates exploration depth based on sample uncertainty, improving both robustness and accuracy.

\subsection{Ophthalmic VLMs and MLLMs}
Early ophthalmic Vision-Language Models (VLMs) like FLAIR \cite{silva2025foundation}, KeepFIT \cite{wu2024mm,wu2025mm}, RetiZero \cite{wang2024common}, and {EyeCLIP \cite{shi2024eyeclip} mainly support image-text representation learning, but are not inherently generative or instruction-following.} This leads to the development of ophthalmic Multimodal Large Language Models (MLLMs) to bridge this gap\cite{luo2025survey}. {EyecareGPT\cite{li2025eyecaregpt} utilizes adaptive resolution and a dense connector for fine-grained analysis. FundusExpert\cite{liu2025constructing} enables positioning-diagnosis collaboration and generates multi-turn cognitive chains. However, they both rely primarily on supervised tuning without RL-based reasoning optimization, which can encourage logical reasoning and generalization through reward-guided exploration beyond static annotations. They also lack explicit CoT and context-rich multi-image complex reasoning supervision, with FundusExpert further limited to the CFP modality.} Moreover, existing ophthalmic MLLMs generally lack an adaptive reasoning mechanism to regulate the exploration progress based on sample uncertainty.

\section{MM-Retinal-Reason}

MM-Retinal-Reason is an ophthalmic multimodal dataset that provides comprehensive data across a full spectrum of perception and reasoning tasks, featuring explicit reasoning traces aligned with real-world clinical scenarios. The dataset comprises four components: basic reasoning VQA, CoT reasoning trajectories, image-caption pairs, and complex reasoning VQA. The overall curation workflow is shown in Fig.~\ref{fig:dataset_category}. {The validation panel comprises two ophthalmologists with more than 15 years of clinical experience and four others with more than 5 years.}

\subsection{Part 1: Basic Reasoning VQA}
\label{sec:basic_reasoning_VQA}
Due to the lack of large-scale ophthalmic VQA datasets with multiple imaging modalities, we construct this subset from 44 public classification datasets, including 55,224 VQA pairs for Color Fundus Photography (CFP), 34,159 for Fundus Fluorescein Angiography (FFA), and 131,985 for Optical Coherence Tomography (OCT).

For question formulation, we adopt templates from PubMedVision\cite{chen2024huatuovision}, MeCoVQA-Complex\cite{huang2025towards}, and DeepSeek-R1\cite{guo2025deepseek}, including single-answer, multi-answer multiple-choice questions (MCQs), and binary true/false questions. {The templates query either the disease depicted in the image for MCQs or the presence of a condition for binary questions.} {Original dataset labels serve as VQA annotations, and candidate options are designed according to dataset type.} For single-label datasets with common diseases (e.g., LAG \cite{li2019attention}, RetinalOCT-C8 \cite{OCTC82021}), we unify categories into a shared candidate pool to increase question difficulty by expanding choice space. For multi-label datasets with diverse and rare diseases (e.g., RFMiD \cite{pachade2021retinal}, STARE \cite{hoover2000locating}\cite{hoover2003locating}), we retain original labels to preserve complexity. To ensure appropriate candidate option sampling, the label pool design follows the label distribution of original datasets to prevent shortcut learning, such as eliminating rare answers or selecting frequent choices. To ensure data quality, a subset of {50} basic VQA samples is selected {using stratified proportional random sampling} and assessed by two ophthalmologists. The assessment shows that 86.0\% of the sampled pairs are clinically acceptable, {with 90.0\% inter-observer agreement}, supporting the reliability and validity of the basic reasoning VQA. {Disagreements are resolved through consensus discussion, with adjudication by senior ophthalmologists.}

\subsection{Part 2: CoT Reasoning Trajectories}
\label{sec:CoT_data}
To better stimulate the model’s reasoning capabilities, we construct CoT reasoning trajectories for the basic reasoning VQA subset in Section~\ref{sec:basic_reasoning_VQA}. The final subset includes 10,270 CFP, 3,274 FFA, and 2,343 OCT CoT reasoning trajectories. 

Following Vision-R1 \cite{huang2025vision}, {we adopt a two-stage CoT generation pipeline. First, Qwen2.5-VL-72B \cite{bai2025qwen2} converts retinal images into structured textual descriptions following ophthalmology-specific guidelines, such as anatomical region scanning and lesion characterization. Then, DeepSeek-R1 generates CoTs by integrating visual cues with step-by-step reasoning, such as candidate option evaluation and differential reasoning.} This strategy decouples perception from reasoning by leveraging the visual {understanding} capability of Qwen2.5-VL-72B and reasoning capability of DeepSeek-R1. Moreover, leveraging the complementary strengths of different models helps mitigate bias amplification from a single generator.

To ensure the clinical reliability, we adopt rigorous quality control. {Samples with incorrect answers are removed. The remaining trajectories are evaluated by Gemini-2.5-Pro~\cite{comanici2025gemini} on a 5-point scale across four aspects: factual correctness (accuracy of visual findings, reasoning, and diagnosis), completeness (thorough visual feature, case information, and structured reasoning), clinical alignment (logical consistency and clinical relevance), and safety (hallucination control).} Trajectories scoring below 3 are discarded. The entire pipeline and rubrics are developed and verified in consultation with ophthalmologists, and a random subset is further reviewed for quality assurance.
To assess the construction pipeline, {50 CoT trajectories are selected from the CoT subset of MM-Retinal-Reason using stratified proportional random sampling and evaluated by three ophthalmologists using the same four-aspect rubric.} {The inter-observer agreement reaches an ICC of 0.816 (95\%CI, 0.73-0.88) and disagreements are resolved by consensus review and senior adjudication.} Furthermore, the expert evaluation yields an acceptance rate of 88.0\%, supporting the reliability of our evaluation process and the clinical validity of the generated CoTs with expert judgment.

\subsection{Part 3: Image-Caption Pairs}
\label{sec:image_caption_pairs}
The image-caption subset comprises 80,520 pairs, spanning diverse ophthalmic imaging modalities, including CFP, OCT, FFA, Computed Tomography (CT), Slit-Lamp, Ultrasound, and others. To construct this subset, we collect ophthalmic articles from PMC up to June 20, 2025. Compound figures from articles are split by morphological operations or Dab-DETR \cite{liu2022dab}. Subcaptions are parsed with Qwen2.5-VL-72B and aligned to subfigures using spatial rules. The final image-caption pairs are manually checked to ensure data quality.

\subsection{Part 4: Complex Reasoning VQA}
Most medical VQA datasets are limited to single-image and basic reasoning (see Fig~\ref{fig:diff_in_dataset}). Although some explore multi-image or complex reasoning, they rarely integrate multimodal multi-image inputs, heterogeneous patient information, clinically relevant questions, and explicit reasoning trajectories. To bridge this gap, we restructure real-world, clinician-authored PMC case reports into complex reasoning VQA with reasoning trajectories, preserving high clinical fidelity and contextual coherence. Since these samples are grounded in authentic clinical narratives rather than generated from scratch, and the pipeline and prompts are developed and verified with ophthalmologists, the dataset maintains high factual fidelity and clinical reliability. The final corpus comprises 5,027 complex reasoning VQA samples, with 400 reserved as the test set covering diverse diseases. {Samples are partitioned at the article level using PMCIDs, ensuring that all samples derived from the same clinical report are assigned to the same split.}

\begin{itemize}
\item \textbf{Step 1: Structured Complex VQA Generation.} Inspired by \cite{wu2025medcasereasoning}, we develop a tailored pipeline to create structured VQA using o4-mini\cite{o4mini}, leveraging its ability to preserve context fidelity and structural coherence with high efficiency. {Specifically, each report is restructured into four components: (1) \textless \texttt{Presentation \& Question}\textgreater: clinical narrative including patient history, examination findings, and other clinical information, ending with a diagnostic question; (2) \textless \texttt{Image Caption}\textgreater: description of all related figures; (3) \textless \texttt{Reasoning Trace}\textgreater: stepwise differential reasoning based on visual cues and source evidence; and (4) \textless \texttt{Answer}\textgreater: the final diagnosis.} To facilitate subsequent evaluation, image descriptions are temporarily retained in \textless \texttt{Presentation \& Question}\textgreater and then removed in Step 3.

\item \textbf{Step 2: Component-wise Quality Evaluation.} {We evaluate the generated VQA using o4-mini and Gemini-2.5-Pro with a rigorous rubric covering major structural components: 1) Image and Case: information completeness, factual correctness, and absence of answer leakage. 2) Reasoning Process: differential diagnosis, logical coherence, and source adherence. 3) Diagnosis Answer: clinical fidelity and derivability from the reasoning trace.}

\item \textbf{Step 3: Visual Information Refinement.} {After selecting valid samples, we employ o4-mini to remove overlapping text between \textless \texttt{Presentation \& Question}\textgreater and \textless \texttt{Image Caption}\textgreater to prevent textual shortcuts. Then, we refine \textless \texttt{Image Caption}\textgreater with the corresponding image captions from Section~\ref{sec:image_caption_pairs} to preserve integrity.}

\item \textbf{Step 4: Hallucination Evaluation.} {We assess the source fidelity to the original case report using o4-mini and Gemini-2.5-Pro, ensuring that case presentation, image descriptions, diagnostic reasoning, and final diagnosis are traceable to the source report. Samples with hallucinations, inconsistencies, or unsupported statements are excluded. A random subset is further reviewed by ophthalmologists to confirm clinical reliability.} 
\end{itemize}

The reliability of the construction pipeline is examined {using the same protocol in Sec.~\ref{sec:CoT_data}} under the evaluation rubric in Step 2. The evaluation yields an average acceptance rate of 94.0\%, {and inter-observer agreement reaches an ICC of 0.830 (95\% CI, 0.75-0.89)}, confirming the quality of complex reasoning VQA.

Notably, unlike MedCaseReasoning \cite{wu2025medcasereasoning}, which contains only textual QA, our dataset provides multimodal and multi-image ophthalmic VQA samples. This design better supports multimodal understanding and diagnostic reasoning aligned with ophthalmologists’ clinical reasoning patterns.

\subsection{Dataset Characteristics and Task Nature}
MM-Retinal-Reason is curated from 44 public datasets and PubMed Central, encompassing over 100 ophthalmic conditions. As illustrated in Fig.~\ref{fig:dataset_category}, the dataset spans broad disease coverage with a long-tailed distribution, including both common and rare abnormalities. Moreover, MM-Retinal-Reason forms a multi-tiered task structure with increasing difficulty: (i) Level 1 involves single-step binary True/False reasoning; (ii) Level 2 focuses on multi-step closed-set reasoning via MCQs; (iii) Level 3 requires complex integrative reasoning over heterogeneous clinical information and multiple images.

As summarized in Tab.~\ref{tab:dataset_nature}, MM-Retinal-Reason includes four complementary components targeting distinct reasoning capabilities. Image-caption pairs align visual findings with descriptions to strengthen perceptual grounding. CoT trajectories provide multi-step rationales to improve evidence chaining and differential diagnosis. Basic reasoning VQA defines a closed-set diagnostic task to evaluate disease recognition and candidate disease discrimination. Complex reasoning VQA focuses on open-ended case reasoning, integrating heterogeneous clinical contexts with multiple images. Together, these designs reflect diverse tasks and clinical scenarios, supporting interpretable and generalizable model learning and evaluation.

\begin{table}[]
\caption{Summary of the four components in MM-Retinal-Reason.}
\label{tab:dataset_nature}
{\renewcommand{\arraystretch}{1.1}
{\setlength{\tabcolsep}{6.9pt}
\begin{tabular}{cccc}
\Xhline{1.1pt}
\textbf{Component}                                                   & \textbf{Input}                                                   & \textbf{Output}                                              & \textbf{Task Nature}                                                                       \\ \hline
\begin{tabular}[c]{@{}c@{}}Image–caption\\ pairs\end{tabular}        & \begin{tabular}[c]{@{}c@{}}Image\\ + Question\end{tabular}       & Caption                                                      & \begin{tabular}[c]{@{}c@{}}Perceptual\\ grounding\end{tabular}                             \\ \hline
\begin{tabular}[c]{@{}c@{}}CoT reasoning\\ trajectories\end{tabular} & \begin{tabular}[c]{@{}c@{}}Image\\ + Question\end{tabular}       & Reasoning                                                    & \begin{tabular}[c]{@{}c@{}}Logical evidence\\ chaining\end{tabular}                        \\ \hline
\begin{tabular}[c]{@{}c@{}}Basic\\ reasoning VQA\end{tabular}        & \begin{tabular}[c]{@{}c@{}}Single image\\ + MCQ\end{tabular}     & \begin{tabular}[c]{@{}c@{}}Diagnosis\\ (option)\end{tabular} & \begin{tabular}[c]{@{}c@{}}Discriminative\\ diagnosis\end{tabular}                         \\ \hline
\begin{tabular}[c]{@{}c@{}}Complex\\ reasoning VQA\end{tabular}      & \begin{tabular}[c]{@{}c@{}}Multi images\\ + Context\end{tabular} & \begin{tabular}[c]{@{}c@{}}Diagnosis\\ (open)\end{tabular}   & \begin{tabular}[c]{@{}c@{}}Context-conditioned\\ integrative case\\ reasoning\end{tabular} \\ 
\Xhline{1.1pt}
\end{tabular}
}
}
\end{table}

\begin{figure*}[!t]
\centerline{\includegraphics[scale=0.6]{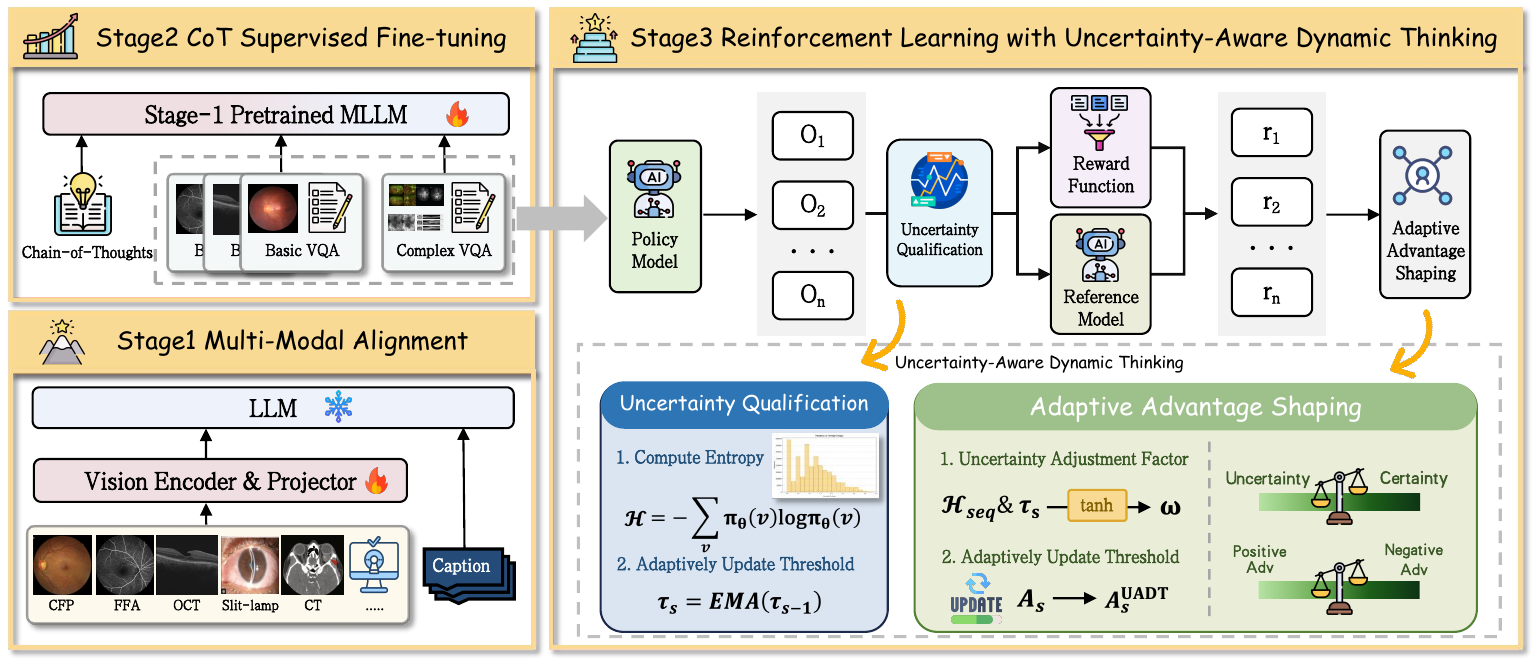}}
\caption{Overview of OphthaReason, which adopts a three-stage training framework: multimodal alignment, CoT supervised fine-tuning, and reinforcement learning. In the third stage, the Uncertainty-Aware Dynamic Thinking mechanism is introduced for adaptive exploration across samples with varying uncertainty levels. OphthaReason demonstrates strong capability in both basic image-based diagnostic reasoning and complex clinical contextualized reasoning.}
\label{fig:model}
\end{figure*}

\section{OphthaReason}
Building upon MM-Retinal-Reason, we introduce OphthaReason, a three-stage multimodal reasoning model designed for full spectrum of ophthalmic diagnosis, from basic image-based reasoning to complex, clinically-simulated reasoning. A key component of OphthaReason is Uncertainty-Aware Dynamic Thinking (UADT), a novel mechanism that enables the model to dynamically adapt exploration depth to tasks of varying difficulty. We developed two versions of OphthaReason for the community, building upon Qwen2.5-VL-3B-Instruct and InternVL3-2B, which are two of the most prominent general-purpose MLLMs with leading reasoning capability. The framework of OphthaReason is illustrated in Fig.~\ref{fig:model}.

\subsection{Stage 1: Foundational Vision-Language Alignment}
While general-purpose MLLMs demonstrate strong foundational vision-language alignment, their capability for interpreting specialized ophthalmic images remains underdeveloped. The text-centric architecture also limits their visual perception \cite{wang2025perception}, weakening the contribution of visual features in reasoning space. To address these issues, we enhance ophthalmic vision-language alignment using the image-caption pair subset from MM-Retinal-Reason and MM-Retinal datasets \cite{wu2024mm}\cite{wu2025mm}. We selectively fine-tune only the vision encoder and projector, leaving the LLM frozen to prevent its advanced language modeling from being degraded by the simple caption data.

\subsection{Stage 2: Activating Reasoning via CoT-SFT}
Prior works \cite{huang2025vision}\cite{tan2025reason}\cite{su2025gmai} have demonstrated that supervised fine-tuning on CoT data effectively unlocks the reasoning abilities of MLLMs through step-by-step logical processes. In the second stage, we fine-tune the whole model using a combination of the reasoning trajectories from both basic and complex reasoning VQA subsets. This preserves the model's structured thinking ability, which would be undermined if solely fine-tuned on final answers. Given a sample consisting of a set of $K$ images $X=\{x_1,x_2\dots,x_K\}$ (where $K\geq1$), a question $q$, a CoT reasoning trajectory $r$, and a final answer $a$, the training objective is to maximize the likelihood of generating the correct reasoning trace and answer:
\begin{align}
\label{eq:sft}
\mathcal{J}_{\text{SFT}}(\theta)=-\mathbb{E}{[(X,q,r,a)\sim\mathcal{P}]}\sum_{t=1}^{T}\text{log}\pi_{\theta}(y_t|X,q,y_{<t}),
\end{align}
where $\mathcal{P}$ represents the dataset, the image number $K=1$ for basic reasoning tasks, and $K\geq1$ for complex reasoning tasks. The term $y$ is the concatenated sequence of reasoning trace $r$ and answer $a$, $t$ indexes the token position in the sequence, and $T$ denotes the total sequence length.

\subsection{Stage 3: Reasoning Enhancement with Uncertainty-Aware Dynamic Thinking RL}
\subsubsection{Policy Optimization}Building on the reasoning-activated model in stage 2, we apply Group Relative Policy Optimization (GRPO) \cite{shao2024deepseekmath} to further enhance the model’s reasoning ability. Given an input sample $(\{x_1,x_2,\dots,x_K\},q,r,a)$ (where $K\geq1$), the policy model $\pi_{\theta}$ generates $G$ candidate outputs $O=\{o_1,o_2,\dots,o_G\}$. Each candidate output $o_i$ is then evaluated by the reward function to yield a corresponding reward $r_i$, forming the reward set $R=\{r_1,r_2,\dots,r_G\}$. Instead of an additional value model, GRPO computes the advantage directly by the following formulation:
\begin{align}
\label{eq:advantage}
A_i=\frac{r_i-\text{mean}(\{r_1,r_2,\dots,r_G\})}{\text{std}(\{r_1,r_2,\dots,r_G\})}.
\end{align}
Then GRPO optimizes the policy model by maximizing the following objective:
\begin{equation}
\label{eq:optimization_objective}
\begin{split}
\mathcal{J}_{\text{GRPO}}(\theta) & =\mathbb{E}{[(X,q,a)\sim\mathcal{P},\{o_i\}_{i=1}^{G}\sim\pi_{\theta_{old}}(O|X,q)}] \\
& \frac{1}{G}\sum_{i=1}^{G}\Big\{\text{min}\Big[\frac{\pi_{\theta}(o_{i}|X,q)}{\pi_{\theta_{old}}(o_{i}|X,q)}A_i^{\text{UADT}}, \\
&\text{clip}\Big(\frac{\pi_{\theta}(o_{i}|X,q)}{\pi_{\theta_{old}}(o_{i}|X,q)},1-\varepsilon,1+\varepsilon\Big)A_i^{\text{UADT}}\Big]\\
&-\beta D_{KL}[\pi_{\theta}||\pi_{\text{ref}}]\Big\},
\end{split}
\end{equation}
where $A_i^{\text{UADT}}$ is the advantage shaped by Uncertainty-Aware Dynamic Thinking (UADT) described in Sec.~\ref{sec:UADT}, which adaptively adjusts the magnitude of the original advantage $A_i$ from Eq.~\ref{eq:advantage}. The hyperparameters $\varepsilon$ and $\beta$ set the clipping range and regulate the strength of the KL divergence penalty, respectively. The KL divergence serves as a regularization term that constrains the policy model $\pi_{\theta}$, preventing excessive deviation from the original policy $\pi_{\text{ref}}$.

\subsubsection{Reward Function}
We employ a composite reward function for each task, comprising a format reward and an accuracy reward, with specific criteria tailored to the task type.
\begin{itemize}
\item \textbf{Format Reward.} This is a binary score for structural adherence. For all tasks, it verifies whether reasoning is encapsulated in \textless \texttt{think} \textgreater \textless \texttt{/think} \textgreater tags, followed by the answer in \textless \texttt{answer}\textgreater \textless \texttt{/answer} \textgreater tags. For open-ended questions, it additionally requires the model to first generate an image description within \textless \texttt{caption} \textgreater \textless \texttt{/caption} \textgreater tags to encourage stronger attention on visual content in complex reasoning tasks.
\item \textbf{Accuracy Reward.} This score measures content correctness. For MCQ task, we compute an IoU score $R_{IoU}=(y\cap\hat{y})/(y\cup\hat{y})$ between the predicted answer $\hat{y}$ and the ground-truth answer $y$, which offers a robust measure for partially correct responses that a simple exact-match would fail. For open-ended questions, the reward is calculated as the weighted score of the model’s Top-5 answers, which are averaged across BLEU, ROUGE, and METEOR to simultaneously evaluate precision, recall, and synonym-aware alignment.
\end{itemize}

\subsection{Uncertainty-Aware Dynamic Thinking}
\label{sec:UADT}
Vanilla GRPO treats all samples uniformly regardless of task difficulty, thus lacking a dynamic mechanism to adapt its exploratory behavior to varying levels of diagnosis case complexity. When confronted with a mixture of basic and complex reasoning problems, this one-fits-all strategy can be suboptimal, leading to insufficient exploration for complex problems and excessive optimization for simpler ones. 

Thus, we propose Uncertainty-Aware Dynamic Thinking (UADT), a novel advantage-shaping mechanism from an entropy perspective, leveraging sample-level uncertainty to regulate the learning process dynamically. For challenging, high-uncertainty tasks, it introduces an entropy bonus to promote exploration. Conversely, for simpler, low-uncertainty tasks, the policy model is encouraged for refinement and stabilization.

\subsubsection{Quantifying Uncertainty via Policy Entropy}
The intrinsic uncertainty of a policy model provides a potent proxy for task difficulty relative to the model's capability, which can be quantified on the fly by the entropy of its output distribution \cite{cheng2025reasoning}\cite{kadavath2022language}. This means that high policy entropy suggests the model is struggling or exploring, indicative of a complex task that is beyond the model's current ability. In contrast, the low policy entropy represents that the model is confident, signifying a simpler task within the model's proficiency.

First, we compute the entropy of the model's prediction on the current sample. For each token position $t$ in the token sequence $Y=\{y_1,y_2,\dots,y_T\}$, the entropy of the predictive distribution over {the model’s tokenizer vocabulary $V$} is calculated by the following equation:
\begin{equation}
\begin{split}
\mathcal{H}_t(y_t | X, q, y_{<t}) = &- \sum_{v \in V} \pi_{\theta}(y_t=v | X, q, y_{<t}) \\
    & \cdot \log \pi_{\theta}(y_t=v | X, q, y_{<t}).
\end{split}
\end{equation}
We then average these per-token entropies across the entire sequence to obtain the final sentence entropy:
\begin{align}
\mathcal{H}_{\text{seq}}(Y | X, q) = \frac{1}{T} \sum_{t=1}^{T} \mathcal{H}_t(y_t | X, q, y_{<t}),
\end{align}
where $X$ is the input image set, and $\pi_{\theta}$ is the policy model.

Next, we measure the model's uncertainty for each sample by comparing its sequence entropy to a predefined threshold, which distinguishes between low and high uncertainty. A fixed threshold is suboptimal as it fails to adapt to the model's evolving reasoning capabilities during training. Thus, we use an adaptive threshold, updated as a moving average of recent entropies, representing the model's average uncertainty on recent samples. Specifically, we initialize the threshold by computing the median entropy of training samples using the base model in inference mode. During training, it is updated using the Exponential Moving Average (EMA) formula:
\begin{align}
\tau_s = \alpha \cdot \tau_{s-1} + (1 - \alpha) \cdot \bar{\mathcal{H}}_s,
\end{align}
where $\tau_s$ and $\tau_{s-1}$ are adaptive thresholds at the current and previous steps, with $\tau_0$ being the initial value. Hyperparameter $\alpha$ denotes the EMA decay factor (set to 0.99 empirically), and $\bar{\mathcal{H}}_s$ indicates the average sequence entropy $\mathcal{H}_{\text{seq}}$ across the samples in the current batch.

\subsubsection{Adaptive Advantage Shaping}
The relationship between the sample-level entropy and adaptive threshold is then mapped to an uncertainty adjustment factor. Instead of a linear function, we utilize $tanh$ activation, which introduces beneficial non-linearity, and normalizes the deviation between the sample's entropy and the threshold. This normalization mitigates the influence of entropy deviations, preventing them from destabilizing the training process. The uncertainty adjustment factor for sample $i$ is defined as:
\begin{equation} \label{eq:difficulty_factor}
\omega_{i,s} = \tanh\left(\gamma \cdot (\mathcal{H}_{seq,i,s} - \tau_s)\right),
\end{equation}
where $\gamma$ is the scaling hyperparameter that controls the steepness of the function.

The factor effectively transforms the sample uncertainty into a continuous signal bounded within $[-1,1]$. 
We then leverage this factor to dynamically modulate the advantage. The final shaped advantage for sample $i$ is as follows:
\begin{equation}
\label{eq:shaped_advantage}
A^{\text{UADT}}_{i,s} = A_{i,s} + \lambda \cdot \omega_{i,s} \cdot \mathcal{H}_{seq,i,s},
\end{equation}
where $A_{i,s}$ stands for the original advantage, and $\lambda$ is a scaling hyperparameter. This shaped advantage is then used to compute the policy optimization objective defined in Eq.~\ref{eq:optimization_objective}, thereby achieving a dynamic, uncertainty-aware policy update.

\begin{table*}[t]
\centering
\caption{\textbf{Results on MM-Retinal-Reason basic reasoning subset of CFP and FFA modalities (\%).} MM denotes MM-Retinal-Reason dataset. S-MCQ and M-MCQ denote Single-answer Multiple-Choice Questions and Multi-answer Multiple-Choice Questions. $^{\dagger}$ The overlapping data between the MM-Retinal-Reason test split and the FundusExpert training set are removed. {$^{\ddagger}$ Following the original EyeCLIP setting, the EyeCLIP vision encoder is connected to Vicuna (Llama 2-7B) for VQA evaluation.} The best results are highlighted in \textbf{Bold}.}
\label{tab:CFP&FFA}

\renewcommand{\arraystretch}{1.12}
\setlength{\tabcolsep}{3.3pt}

\renewcommand{\wci}[3]{#1\,\raisebox{0.3ex}{\tiny{[#2,#3]}}}

\begin{adjustbox}{width=\textwidth}
\begin{tabular}{ccc ccccccccc ccc}
\toprule
\multirow[c]{5}{*}{\makecell[c]{\textbf{Category}}}
& \multirow[c]{5}{*}{\makecell[c]{\textbf{Model}}}
& \multirow[c]{5}{*}{\makecell[c]{\textbf{Params}}} 
& \multicolumn{9}{c}{\textbf{MM-CFP}} 
& \multicolumn{3}{c}{\textbf{MM-FFA}} \\
\cmidrule(lr){4-12} \cmidrule(lr){13-15}

& & 
& \multicolumn{1}{c}{\textbf{ID}} 
& \multicolumn{7}{c}{\textbf{OOD}} 
& \multirow{3}{*}{\textbf{AVG {(95\%CI)}}} 
& \multirow{2}{*}{\textbf{ID}} 
& \multirow{2}{*}{\textbf{OOD}} 
& \multirow{3}{*}{\textbf{AVG {(95\%CI)}}} \\

& & 
& \multicolumn{1}{c}{\textbf{S-MCQ}} 
& \multicolumn{1}{c}{\textbf{S-MCQ}} 
& \multicolumn{6}{c}{\textbf{M-MCQ}} 
& 
& 
& 
& \\
\cmidrule(lr){4-4} \cmidrule(lr){5-11} \cmidrule(lr){13-14}

& & 
& \textbf{Acc} 
& \textbf{Acc} 
& \textbf{Jaccard} 
& \textbf{Acc} 
& \textbf{Precision} 
& \textbf{Recall} 
& \textbf{F1} 
& \textbf{AVG} 
& 
& \textbf{Acc} 
& \textbf{Acc} 
& \\
\midrule

\multirow{3}{*}{\makecell[c]{Proprietary\\General Models}}
& GPT-4o\cite{achiam2023gpt} 
& -- 
& 59.50 & 62.06 & 45.81 & 30.79 & 53.61 & 54.41 & 51.69 & 47.26 & \wci{56.27}{55.24}{57.33} & 43.75 & 55.50 & \wci{49.63}{46.25}{53.00} \\
& GPT-4V(ision)\cite{gpt4V} 
& -- 
& 46.14 & 43.82 & 31.35 & 22.38 & 35.80 & 36.74 & 34.90 & 32.23 & \wci{40.73}{39.67}{41.79} & 36.00 & 47.50 & \wci{41.75}{38.38}{45.13} \\
& Gemini-2.5-Flash-Lite\cite{comanici2025gemini} 
& -- 
& 41.53 & 41.26 & 33.52 & 18.48 & 40.43 & 42.98 & 39.36 & 34.95 & \wci{39.25}{38.24}{40.25} & 35.00 & 54.00 & \wci{44.50}{41.25}{47.88} \\
\midrule

\multirow{4}{*}{\makecell[c]{Open-Source\\General Models}}
& LLaVA-Next\cite{li2024llava} 
& 8B 
& 17.18 & 24.59 & 24.05 & 12.09 & 31.29 & 34.53 & 29.26 & 26.24 & \wci{22.67}{21.82}{23.54} & 26.25 & 38.25 & \wci{32.25}{29.13}{35.38} \\
& Qwen2.5-VL-Instruct\cite{bai2025qwen2} 
& 3B 
& 31.07 & 32.62 & 26.21 & 13.59 & 34.87 & 32.90 & 31.21 & 27.76 & \wci{30.48}{29.52}{31.48} & 29.75 & 29.00 & \wci{29.38}{26.25}{32.63} \\
& Qwen2.5-VL-Instruct\cite{bai2025qwen2} 
& 7B 
& 46.65 & 41.86 & 29.92 & 13.30 & 37.45 & 42.08 & 36.63 & 31.88 & \wci{40.13}{39.13}{41.14} & 37.75 & 38.00 & \wci{37.88}{34.50}{41.25} \\
& InternVL-3\cite{chen2024internvl} 
& 2B 
& 54.42 & 52.35 & 27.52 & 15.27 & 32.19 & 37.18 & 32.37 & 28.91 & \wci{45.23}{44.22}{46.23} & 26.00 & 44.50 & \wci{35.25}{32.00}{38.50} \\
\midrule

\multirow{6}{*}{\makecell[c]{General Medical\\Models w/o RL}}
& LLaVA-Med\cite{li2023llava} 
& 7B 
& 31.29 & 33.22 & 25.88 & 19.07 & 34.88 & 25.88 & 28.48 & 26.84 & \wci{30.45}{28.24}{32.62} & 26.55 & 28.95 & \wci{27.75}{22.42}{33.08} \\
& Med-Flamingo\cite{moor2023med} 
& 9B 
& 28.51 & 28.86 & 22.15 & 15.09 & 32.47 & 22.15 & 24.98 & 23.37 & \wci{26.91}{25.94}{27.85} & 20.25 & 30.50 & \wci{25.38}{22.38}{28.38} \\
& HealthGPT\cite{lin2025healthgpt} 
& 3.8B 
& 42.36 & 38.08 & 32.24 & 18.42 & 38.22 & 47.16 & 38.40 & 34.89 & \wci{38.44}{37.44}{39.45} & 26.75 & 32.00 & \wci{29.38}{26.25}{32.50} \\
& HuatuoGPT-Vision\cite{chen2024huatuogpt} 
& 7B 
& 57.38 & 50.22 & 38.07 & 28.25 & 50.46 & 38.32 & 41.75 & 39.37 & \wci{48.99}{47.94}{50.02} & 39.25 & 45.75 & \wci{42.50}{39.13}{45.88} \\
& Lingshu\cite{xu2025lingshu} 
& 7B 
& 71.28 & 63.38 & 41.44 & 30.22 & 52.92 & 44.20 & 45.86 & 42.93 & \wci{59.20}{58.19}{60.19} & 36.25 & 40.75 & \wci{38.50}{35.13}{41.88} \\
& Hulu-Med\cite{jiang2025hulu} 
& 4B 
& 65.76 & 58.30 & 40.19 & 28.93 & 51.70 & 42.90 & 44.59 & 41.66 & \wci{55.24}{54.22}{56.24} & 30.75 & 45.75 & \wci{38.25}{35.00}{41.50} \\
\midrule

\multirow{3}{*}{\makecell[c]{General Medical\\Models w/ RL}}
& MedVLM-R1\cite{pan2025medvlm} 
& 2B 
& 31.73 & 30.57 & 27.70 & 5.44 & 32.92 & 48.12 & 36.54 & 30.14 & \wci{30.81}{29.89}{31.73} & 27.00 & 26.00 & \wci{26.50}{23.50}{29.50} \\
& QoQ-Med\cite{dai2025qoq} 
& 7B 
& 59.63 & 51.95 & 34.33 & 16.74 & 39.81 & 48.93 & 41.40 & 36.24 & \wci{49.27}{48.26}{50.30} & 40.50 & 32.00 & \wci{36.25}{33.00}{39.50} \\
& {MedGemma\cite{sellergren2025medgemma}}
& 4B 
& 72.10 & 59.30 & 37.79 & 19.74 & 44.01 & 51.05 & 44.64 & 39.45 & \wci{56.95}{55.97}{57.90} & 34.50 & 34.50 & \wci{34.50}{31.25}{37.75} \\
\midrule

\multirow{6}{*}{{\makecell[c]{Ophthalmic\\Models}}}
& FundusExpert$^{\dagger}$\cite{liu2025constructing} 
& 8B 
& 77.89 & 70.86 & 48.72 & 38.68 & 60.94 & 52.52 & 53.02 & \textbf{50.78} & \wci{66.51}{65.37}{67.64} & 29.50 & 49.25 & \wci{39.38}{36.13}{42.75} \\
& Med-R1 (CFP)\cite{lai2025med} 
& 2B 
& 38.17 & 35.78 & 32.23 & 10.98 & 37.62 & 51.68 & 40.91 & 34.68 & \wci{36.21}{35.24}{37.18} & 26.75 & 32.25 & \wci{29.50}{26.38}{32.75} \\
& {EyeCLIP$^{\ddagger}$\cite{shi2024eyeclip}}
& 7B
& 60.14 & 51.62 & 31.66 & 24.79 & 41.46 & 31.69 & 34.37 & 32.79 & \wci{48.18}{47.15}{49.23} & 39.00 & 21.00 & \wci{30.00}{27.00}{33.13} \\
& {EyecareGPT\cite{li2025eyecaregpt}}
& 3.8B
& 67.65 & 61.14 & 34.68 & 15.24 & 41.25 & \textbf{53.19} & 42.54 & 37.38 & \wci{55.39}{54.39}{56.37} & 45.00 & 51.50 & \wci{48.25}{44.88}{51.75} \\
& \cellcolor[HTML]{E0FFFF}OphthaReason-Qwen 
& \cellcolor[HTML]{E0FFFF}3B 
& \cellcolor[HTML]{E0FFFF}74.25 & \cellcolor[HTML]{E0FFFF}71.08 & \cellcolor[HTML]{E0FFFF}42.11 & \cellcolor[HTML]{E0FFFF}31.01 & \cellcolor[HTML]{E0FFFF}51.42 & \cellcolor[HTML]{E0FFFF}46.30 & \cellcolor[HTML]{E0FFFF}46.46 & \cellcolor[HTML]{E0FFFF}43.46 & \cellcolor[HTML]{E0FFFF}\wci{62.93}{61.97}{63.88} & \cellcolor[HTML]{E0FFFF}64.75 & \cellcolor[HTML]{E0FFFF}45.00 & \cellcolor[HTML]{E0FFFF}\wci{54.88}{51.50}{58.38} \\
& \cellcolor[HTML]{E0FFFF}OphthaReason-Intern 
& \cellcolor[HTML]{E0FFFF}2B 
& \cellcolor[HTML]{E0FFFF}\textbf{83.75} & \cellcolor[HTML]{E0FFFF}\textbf{73.78} & \cellcolor[HTML]{E0FFFF}\textbf{49.48} & \cellcolor[HTML]{E0FFFF}\textbf{38.70} & \cellcolor[HTML]{E0FFFF}\textbf{61.33} & \cellcolor[HTML]{E0FFFF}50.76 & \cellcolor[HTML]{E0FFFF}\textbf{53.52} & \cellcolor[HTML]{E0FFFF}50.76 & \cellcolor[HTML]{E0FFFF}\wci{\textbf{69.43}}{68.53}{70.34} & \cellcolor[HTML]{E0FFFF}\textbf{74.25} & \cellcolor[HTML]{E0FFFF}\textbf{61.75} & \cellcolor[HTML]{E0FFFF}\wci{\textbf{68.00}}{64.75}{71.13} \\
\bottomrule
\end{tabular}
\end{adjustbox}
\end{table*}

\subsubsection{Discussion and Analysis}
UADT modulates the policy update by leveraging entropy to gauge sample-level uncertainty and advantage to assess action quality. We analyze its behavior across four scenarios, each defined by a combination of entropy and advantage:
\begin{itemize}
\item \textbf{High Uncertainty \& Positive Advantage. }The model takes a beneficial action but is uncertain ($\mathcal{H}_{seq,i,s}>\tau_s, A_{i,s}>0$). UADT amplifies the shaped advantage, acting as a bonus to encourage exploration.
\item \textbf{High Uncertainty \& Negative Advantage. }The model takes a negative action under high uncertainty ($\mathcal{H}_{seq,i,s}>\tau_s, A_{i,s}<0$) . The positive uncertainty factor promotes exploration of challenging instances, preventing premature convergence, similar to hard sample mining.
\item \textbf{Low Uncertainty \& Positive Advantage. }The model is confident and correct ($\mathcal{H}_{seq,i,s}<\tau_s, A_{i,s}>0$). The shaped advantage is slightly suppressed, preventing over-optimization on mastered samples and shifting focus to more uncertain tasks.
\item \textbf{Low Uncertainty \& Negative Advantage. }This scenario shows an overconfident mistake, where the model is highly confident but the action is wrong ($\mathcal{H}_{seq,i,s}<\tau_s, A_{i,s}<0$). UADT punishes these errors, encouraging mistake correction and enhancing model reliability.
\end{itemize}

UADT introduces an advanced and effective approach, offering several key strengths:
\begin{itemize}
\item \textbf{Adaptive Exploration.} It is critical for datasets with varying difficulty. UADT intelligently allocates its exploration budget to mitigate dual risks of under-exploration on complex tasks and over-exploration on simple ones.
\item {\textbf{Automated and Stable Adaptation.} UADT leverages intrinsic model entropy without external supervision. Its EMA-based threshold and tanh normalization ensure robustness against evolving model proficiency, preventing abrupt policy updates. The training process also exhibits stable optimization and low hyperparameter sensitivity.}
\item {\textbf{Computational Efficiency and Practicality.} UADT is a lightweight, plug-and-play mechanism that requires only minor code modifications. It introduces negligible training overhead with minimal additional memory and time cost, making it well-suited for deployment in resource-constrained environments.}

\end{itemize}

{Unlike entropy-regularized and entropy-control methods\cite{haarnoja2018soft}\cite{gaidemystifying}\cite{shen2026entropy}, UADT does not optimize a differentiable entropy objective, but uses response entropy relative to a temporal reference to adjust the GRPO advantage. Compared with state-level exploration\cite{han2021max} and one-sided token-level modulation\cite{cheng2025reasoning}\cite{liu2026stable}, UADT performs bidirectional response-level regulation using an evolving temporal reference.}

\section{Experiments}

\subsection{Data and Experimental Setup}
\subsubsection{Training Data}
{OphthaReason is trained under a strict train–test separation protocol throughout all stages to prevent data leakage. Specifically, training and test sets contain disjoint samples, and all data derived from the same source are assigned to the same split.} In the initial multimodal alignment stage, the model is trained on 9,437 samples from MM-Retinal~\cite{wu2024mm}\cite{wu2025mm} and 72,861 samples from the image-caption training set of MM-Retinal-Reason, with 7,659 samples overlapping with the complex reasoning VQA test set explicitly excluded. Consistently, the second and third training stages use 20,514 CoT and 28,438 VQA samples from the corresponding training splits of MM-Retinal-Reason.

\subsubsection{Evaluation Data}
For comprehensive evaluation, we assess the model on the MM-Retinal-Reason test set and {three} external datasets. {MM-Retinal-Reason test set follows a strict dataset-level split}, including both in-distribution (ID) test data from {24 datasets} and {out-of-domain (OOD) data from 20 held-out datasets never seen during training}. Each component dataset contributes up to 400 cases while preserving its original disease distribution, with moderate downsampling of healthy-dominant datasets to improve disease coverage. External evaluation assesses basic reasoning on the fundus and OCT branches of GMAI-MMBench \cite{ye2024gmai} and {the ophthalmology-specific LMOD benchmark\cite{qin2025lmod} after excluding samples overlapping with the MM-Retinal-Reason training split}. External complex reasoning is evaluated on the ophthalmology-related samples of MedXpertQA-MM \cite{zuo2025medxpertqa}. {These evaluation datasets encompass diverse institutional sources and collections across at least 14 countries, heterogeneous patient populations and imaging systems, and over 100 disease and finding categories.}

\subsubsection{Comparison Models}
To validate OphthaReason's effectiveness and generalizability, we compare it against 20 models across basic and complex reasoning tasks, including proprietary and open-source general-purpose MLLMs, general medical MLLMs with and without RL, {and ophthalmic models. The ophthalmic baselines include FundusExpert, the fundus and OCT versions of Med-R1, EyecareGPT, and an EyeCLIP-based VQA model constructed by integrating the EyeCLIP vision encoder with Vicuna (Llama 2-7B) following the original EyeCLIP study.} 

\subsubsection{Implementation Details}
In the initial vision-language alignment stage, Qwen-based variant and Intern-based variant are trained for one epoch at learning rates of $2e-6$ and $2e-5$. For the CoT supervision stage, variants are trained for two epochs with learning rates adjusted to $1e-6$ and $1e-5$. In the RL stage, both variants are trained {via full-parameter tuning} for two epochs using a learning rate of $1e-6$. We use AdamW as the optimizer throughout all training stages, and the sampling temperature is 1.0. The UADT hyperparameters $\gamma$ and $\lambda$ are set to 1.0 and 0.5. The main training is conducted on 6 NVIDIA RTX A6000 GPUs, while the 9B-scale experiment is trained on 6 NVIDIA H100 GPUs. All evaluations are performed using the vLLM framework\cite{kwon2023efficient}.

\subsubsection{Statistical Analysis}
{We report 95\% confidence intervals (CIs) for AVG scores using stratified bootstrap with 10,000 replicates and assess statistical significance using two-sided paired permutation tests with 10,000 permutations.}

\begin{table*}[t]
\centering
\caption{\textbf{Results on MM-Retinal-Reason basic reasoning subset of OCT modality and complex reasoning subset(\%).} MM denotes MM-Retinal-Reason dataset. {$^{\ddagger}$ Following the original EyeCLIP VQA setting, the EyeCLIP vision encoder is integrated with Vicuna (Llama 2-7B).} The best results among open-source models are highlighted in \textbf{Bold} (results of large-scale proprietary general models are only reported for reference).}
\label{tab:OCT&complex}

\renewcommand{\arraystretch}{1.12}
\setlength{\tabcolsep}{3.6pt}

\begin{adjustbox}{width=\textwidth}
\begin{tabular}{ccc ccc ccccccc}
\toprule
\multirow[c]{4}{*}{\makecell[c]{\textbf{Category}}}
& \multirow[c]{4}{*}{\makecell[c]{\textbf{Model}}}
& \multirow[c]{4}{*}{\makecell[c]{\textbf{Params}}} 
& \multicolumn{3}{c}{\textbf{MM-OCT}}
& \multicolumn{7}{c}{\textbf{MM-Complex Reasoning}} \\
\cmidrule(lr){4-6} \cmidrule(lr){7-13}
& & 
& \makecell[c]{\textbf{ID}\\\textbf{Acc}}
& \makecell[c]{\textbf{OOD}\\\textbf{Acc}}
& \textbf{AVG {(95\%CI)}}
& \textbf{Recall}
& \textbf{BLEU}
& \textbf{ROUGE-1}
& \textbf{ROUGE-L}
& \textbf{METEOR}
& \textbf{BERTScore}
& \textbf{AVG {(95\%CI)}} \\
\midrule

\multirow{3}{*}{\makecell[c]{Proprietary\\General Models}}
& GPT-4o\cite{achiam2023gpt}
& --
& 34.48 & 52.63 & 43.56 {\raisebox{0.3ex}{\tiny [41.38, 45.73]}}
& 50.82 & 39.58 & 49.69 & 49.40 & 42.65 & 44.64 & 46.13 {\raisebox{0.3ex}{\tiny [42.40, 49.93]}} \\
& GPT-4V(ision)\cite{gpt4V}
& --
& 29.89 & 45.75 & 37.82 {\raisebox{0.3ex}{\tiny [35.68, 39.97]}}
& 43.71 & 35.48 & 44.35 & 43.74 & 36.68 & 41.31 & 40.88 {\raisebox{0.3ex}{\tiny [37.07, 44.68]}} \\
& Gemini-2.5-Flash-Lite\cite{comanici2025gemini}
& --
& 30.78 & 49.06 & 39.92 {\raisebox{0.3ex}{\tiny [37.80, 42.00]}}
& 39.28 & 35.95 & 40.98 & 40.88 & 32.76 & 39.47 & 38.22 {\raisebox{0.3ex}{\tiny [34.35, 42.11]}} \\
\midrule

\multirow{4}{*}{\makecell[c]{Open-Source\\General Models}}
& LLaVA-Next\cite{li2024llava}
& 8B
& 29.22 & 30.75 & 29.99 {\raisebox{0.3ex}{\tiny [28.01, 31.97]}}
& 12.10 & 10.18 & 14.04 & 14.04 & 9.39 & 10.67 & 11.74 {\raisebox{0.3ex}{\tiny [9.54, 14.03]}} \\
& Qwen2.5-VL-Instruct\cite{bai2025qwen2}
& 3B
& 28.94 & 40.13 & 34.54 {\raisebox{0.3ex}{\tiny [32.46, 36.56]}}
& 12.03 & 9.11 & 12.90 & 12.90 & 8.89 & 11.10 & 11.16 {\raisebox{0.3ex}{\tiny [8.92, 13.48]}} \\
& Qwen2.5-VL-Instruct\cite{bai2025qwen2}
& 7B
& 28.73 & 38.50 & 33.62 {\raisebox{0.3ex}{\tiny [31.51, 35.63]}}
& 17.93 & 13.94 & 18.97 & 18.90 & 13.07 & 16.51 & 16.55 {\raisebox{0.3ex}{\tiny [13.94, 19.28]}} \\
& InternVL3\cite{chen2024internvl}
& 2B
& 30.61 & 32.13 & 31.37 {\raisebox{0.3ex}{\tiny [29.35, 33.39]}}
& 16.70 & 13.26 & 17.91 & 17.83 & 12.46 & 15.00 & 15.53 {\raisebox{0.3ex}{\tiny [12.96, 18.20]}} \\
\midrule

\multirow{6}{*}{\makecell[c]{General Medical\\Models w/o RL}}
& LLaVA-Med\cite{li2023llava}
& 7B
& 28.09 & 27.85 & 27.97 {\raisebox{0.3ex}{\tiny [25.04, 30.92]}}
& 18.88 & 4.74 & 9.27 & 8.92 & 12.13 & 7.22 & 10.19 {\raisebox{0.3ex}{\tiny [8.92, 11.54]}} \\
& Med-Flamingo\cite{moor2023med}
& 9B
& 29.15 & 33.25 & 31.20 {\raisebox{0.3ex}{\tiny [30.14, 34.18]}}
& 13.50 & 11.25 & 15.42 & 15.44 & 9.35 & 12.98 & 12.98 {\raisebox{0.3ex}{\tiny [10.74, 15.33]}} \\
& HealthGPT\cite{lin2025healthgpt}
& 3.8B
& 30.75 & 32.63 & 31.69 {\raisebox{0.3ex}{\tiny [29.63, 33.73]}}
& 20.77 & 17.61 & 22.97 & 22.87 & 16.15 & 20.60 & 20.16 {\raisebox{0.3ex}{\tiny [17.21, 23.20]}} \\
& HuatuoGPT-Vision\cite{chen2024huatuogpt}
& 7B
& 41.42 & 53.63 & 47.53 {\raisebox{0.3ex}{\tiny [45.34, 49.72]}}
& 21.84 & 13.80 & 20.78 & 20.56 & 16.63 & 15.59 & 18.20 {\raisebox{0.3ex}{\tiny [15.69, 20.76]}} \\
& Lingshu\cite{xu2025lingshu}
& 7B
& 39.75 & 49.13 & 44.44 {\raisebox{0.3ex}{\tiny [42.26, 46.58]}}
& \textbf{26.67} & 23.43 & \textbf{28.30} & \textbf{28.12} & \textbf{21.31} & 26.24 & \textbf{25.68} {\raisebox{0.3ex}{\tiny [22.49, 29.04]}} \\
& Hulu-Med\cite{jiang2025hulu}
& 4B
& 53.83 & 53.25 & 53.54 {\raisebox{0.3ex}{\tiny [51.42, 55.66]}}
& 24.76 & 20.98 & 26.24 & 26.22 & 19.30 & 22.82 & 23.39 {\raisebox{0.3ex}{\tiny [20.20, 26.60]}} \\
\midrule

\multirow{3}{*}{\makecell[c]{General Medical\\Models w/ RL}}
& MedVLM-R1\cite{pan2025medvlm}
& 2B
& 27.27 & 33.75 & 30.51 {\raisebox{0.3ex}{\tiny [28.54, 32.49]}}
& 12.36 & 9.90 & 13.52 & 13.52 & 9.08 & 11.68 & 11.68 {\raisebox{0.3ex}{\tiny [9.55, 13.94]}} \\
& QoQ-Med\cite{dai2025qoq}
& 7B
& 30.47 & 41.30 & 35.89 {\raisebox{0.3ex}{\tiny [33.81, 37.95]}}
& 22.38 & 19.24 & 24.47 & 24.44 & 17.45 & 20.87 & 21.48 {\raisebox{0.3ex}{\tiny [18.52, 24.61]}} \\
& {MedGemma\cite{sellergren2025medgemma}}
& 4B
& 31.52 & 36.75 & 34.14 {\raisebox{0.3ex}{\tiny [32.08, 36.20]}}
& 22.52 & 18.65 & 23.70 & 23.63 & 17.79 & 20.60 & 21.15 {\raisebox{0.3ex}{\tiny [18.15, 24.27]}} \\
\midrule

\multirow{6}{*}{{\makecell[c]{Ophthalmic\\Models}}}
& FundusExpert\cite{liu2025constructing}
& 8B
& 31.45 & 45.75 & 38.60 {\raisebox{0.3ex}{\tiny [36.46, 40.69]}}
& 13.09 & 10.96 & 15.12 & 15.05 & 9.85 & 12.16 & 12.71 {\raisebox{0.3ex}{\tiny [10.42, 15.11]}} \\
& Med-R1 (OCT/CFP)\cite{lai2025med}
& 2B
& 28.87 & 26.88 & 27.88 {\raisebox{0.3ex}{\tiny [25.95, 29.80]}}
& 10.04 & 8.01 & 11.17 & 11.17 & 7.30 & 9.91 & 9.60 {\raisebox{0.3ex}{\tiny [7.62, 11.76]}} \\
& {EyeCLIP$^{\ddagger}$\cite{shi2024eyeclip}}
& 7B
& 24.27 & 37.50 & 30.89 {\raisebox{0.3ex}{\tiny [28.92, 32.86]}}
& 17.07 & 15.20 & 19.61 & 19.53 & 13.55 & 19.07 & 17.34 {\raisebox{0.3ex}{\tiny [14.70, 20.05]}} \\
& {EyecareGPT\cite{li2025eyecaregpt}}
& 3.8B
& 51.60 & 55.13 & 53.37 {\raisebox{0.3ex}{\tiny [51.18, 55.50]}}
& 24.36 & 16.93 & 23.02 & 22.72 & 18.36 & 20.06 & 20.91 {\raisebox{0.3ex}{\tiny [18.17, 23.72]}} \\
& \cellcolor[HTML]{E0FFFF}OphthaReason-Qwen
& \cellcolor[HTML]{E0FFFF}3B
& \cellcolor[HTML]{E0FFFF}56.07 & \cellcolor[HTML]{E0FFFF}54.63 & \cellcolor[HTML]{E0FFFF}55.35 {\raisebox{0.3ex}{\tiny [53.18, 57.51]}}
& \cellcolor[HTML]{E0FFFF}24.11 & \cellcolor[HTML]{E0FFFF}21.92 & \cellcolor[HTML]{E0FFFF}26.48 & \cellcolor[HTML]{E0FFFF}26.42 & \cellcolor[HTML]{E0FFFF}18.75 & \cellcolor[HTML]{E0FFFF}23.41 & \cellcolor[HTML]{E0FFFF}23.52 {\raisebox{0.3ex}{\tiny [20.49, 26.63]}} \\
& \cellcolor[HTML]{E0FFFF}OphthaReason-Intern
& \cellcolor[HTML]{E0FFFF}2B
& \cellcolor[HTML]{E0FFFF}\textbf{81.87} & \cellcolor[HTML]{E0FFFF}\textbf{57.13} & \cellcolor[HTML]{E0FFFF}\textbf{69.50} {\raisebox{0.3ex}{\tiny [67.48, 71.46]}}
& \cellcolor[HTML]{E0FFFF}25.65 & \cellcolor[HTML]{E0FFFF}\textbf{23.82} & \cellcolor[HTML]{E0FFFF}27.55 & \cellcolor[HTML]{E0FFFF}27.52 & \cellcolor[HTML]{E0FFFF}20.79 & \cellcolor[HTML]{E0FFFF}\textbf{26.75} & \cellcolor[HTML]{E0FFFF}25.35 {\raisebox{0.3ex}{\tiny [22.14, 28.66]}} \\
\bottomrule
\end{tabular}
\end{adjustbox}
\end{table*}

\subsection{Performance on Basic Reasoning Tasks}

We evaluate the basic reasoning performance of OphthaReason under both ID and OOD scenarios by conducting assessments across CFP, FFA, and OCT modalities. For the CFP modality, model performance is evaluated on both single-answer and multiple-answer MCQs. For the FFA and OCT modalities, only single-answer MCQs are included due to the scarcity of suitable public datasets. 

\subsubsection{In-Domain Effectiveness} 
As shown in Tab.~\ref{tab:CFP&FFA} and Tab.~\ref{tab:OCT&complex}, OphthaReason significantly surpasses all competing models. Specifically, OphthaReason-Intern attains an accuracy of 83.75\% for CFP modality, outperforming the second-best approach by 5.86\% {(\textit{P}$<$0.05)}. This edge is also evident in FFA and OCT modalities, where our model exceeds the second-best models by {29.25\%} and 28.04\% {(both \textit{P}$<$0.001)}. This consistent state-of-the-art performance across modalities underscores the effectiveness of OphthaReason, establishing a solid foundation for its generalization and efficiency. 

\subsubsection{Out-of-Domain Generalization}

\begin{figure}[]
\centerline{\includegraphics[scale=0.27]{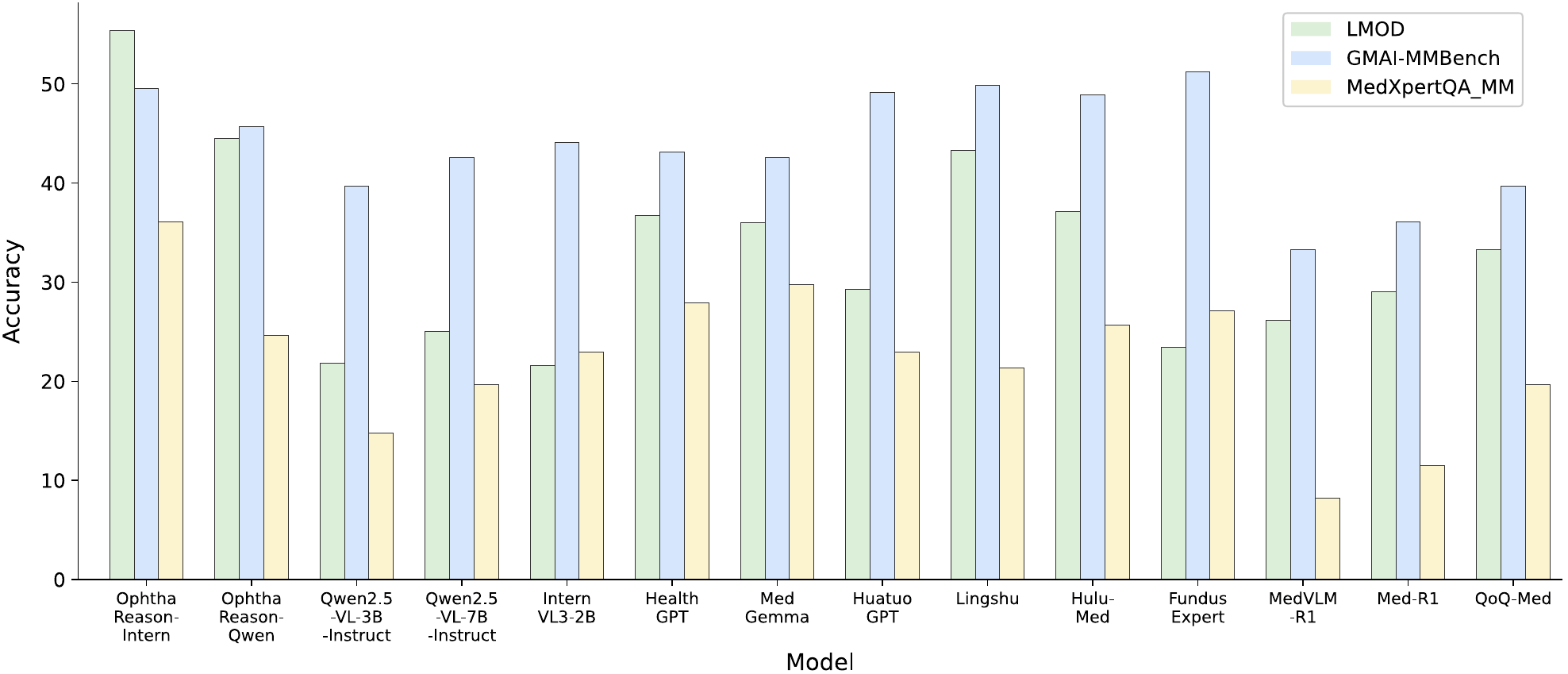}}
\caption{Performance comparison on {LMOD,} GMAI-MMBench (Fundus and OCT subsets), and MedXpertQA-MM (ophthalmology-related samples).}
\label{fig:experiment}
\end{figure}

Results in Tab.~\ref{tab:CFP&FFA} and ~\ref{tab:OCT&complex} also show that OphthaReason achieves strong performance under OOD settings. For CFP, it achieves 73.78\% accuracy on single-answer MCQs and 50.76\% average score on multi-answer MCQs. Notably, results on multi-answer MCQs are comparable to those of FundusExpert, despite OphthaReason using only one-quarter of the parameters (2B vs. 8B), and FundusExpert being more explicitly tailored for the CFP modality. Furthermore, OphthaReason maintains performance margins under OOD settings for both FFA and OCT modalities, demonstrating its robust generalization capability across unseen data.
We further conduct experiments on external {LMOD} and GMAI-MMBench. As illustrated in Fig.~\ref{fig:experiment}, {OphthaReason variants rank first and second on LMOD, achieving accuracies of 55.37\% and 44.51\% and outperforming Lingshu by 12.08\% and 1.22\%.} On GMAI-MMBench, they remain superior performance, competitive with FundusExpert and Lingshu while outperforming most other models, demonstrating generalization beyond MM-Retinal-Reason to external benchmarks.

\subsubsection{Higher Parameter Efficiency}
OphthaReason-Intern-2B and OphthaReason-Qwen-3B are compared against similarly-sized models, much larger 7B-8B models, and massive-scale proprietary models like GPT-4o. The results in Tab.~\ref{tab:CFP&FFA} and Tab.~\ref{tab:OCT&complex} reveal that OphthaReason achieves leading performance in a highly parameter-efficient manner. 
In the CFP modality, its average score of 69.43\% significantly surpasses specialized models like Lingshu-7B (59.20\%, {\textit{P}$<$0.001}), as well as the proprietary GPT-4o (56.27\%, {\textit{P}$<$0.001}) and GPT-4V (40.73\%, {\textit{P}$<$0.001}). Similarly, in the FFA and OCT modalities, OphthaReason surpasses all baselines under both ID and OOD settings, achieving the average scores of 68.00\% and 69.50\% {(all \textit{P}$<$0.001)}.

\begin{figure*}[!t]
\centerline{\includegraphics[scale=0.58]{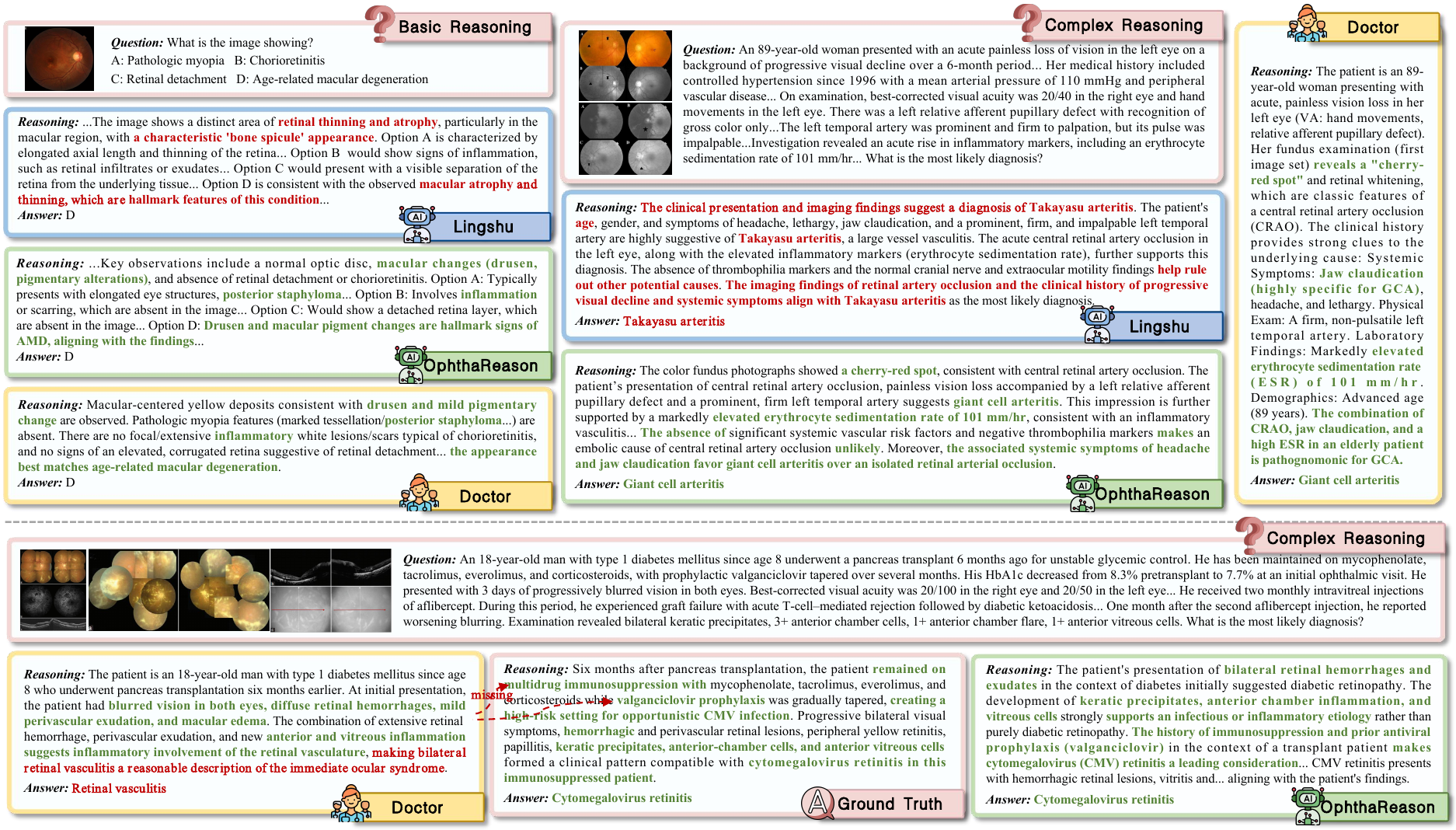}}
\caption{\textbf{Qualitative results of basic reasoning and complex reasoning (upper panel) and {a comparison between ophthalmologist and OphthaReason (lower panel)}.} Red text and arrows indicate omitted key information, factual incorrectness, hallucination, or insufficient reasoning, whereas green text denotes correct and well-structured reasoning. The complex reasoning cases are adapted from \cite{chu2010concurrent}\cite{chao2019bilateral} and are used under the Open Access license.}
\label{fig:case_study}
\end{figure*}

\begin{figure*}[!t]
\centerline{\includegraphics[scale=0.63]{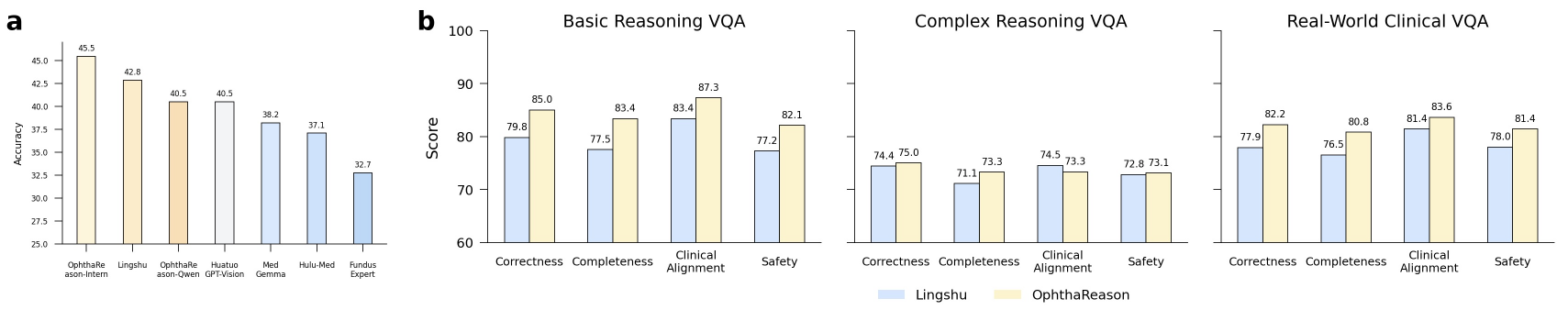}}
\caption{(a) Performance evaluation on real-world clinical reasoning task. (b) Expert evaluation of the reasoning processes for OphthaReason and Lingshu.}
\label{fig:clinical_reasoning}
\end{figure*}

\subsection{Performance on Complex Reasoning Tasks}
In addition to the basic reasoning tasks, we further assess OphthaReason's advanced reasoning capabilities on the MM-Retinal-Reason Complex Reasoning, a challenging task designed to simulate real-world clinical scenarios, requiring models to integrate information from both multimodal images and detailed clinical information. Moreover, we also evaluate our model on the public MedXpertQA-MM, another benchmark for complex reasoning. As it covers a broad range of medical domains, we only conduct the evaluation on ophthalmology-related samples.

\subsubsection{Advanced Integrative Diagnostic Reasoning}
As shown in Tab.~\ref{tab:OCT&complex}, OphthaReason achieves an average score of 25.35\% on answer correctness and semantic alignment metrics, outperforming general medical models like QoQ-Med (+3.87\%, {\textit{P}$<$0.05}) and Hulu-Med (+1.96\%, {\textit{P}$>$0.05}), as well as ophthalmic models such as {EyecareGPT (+4.44\%, \textit{P}$<$0.05), EyeCLIP (+8.01\%, \textit{P}$<$0.001), and FundusExpert (+12.64\%, \textit{P}$<$0.001).} OphthaReason is highly competitive with Lingshu, with only a marginal gap of 0.33\% {(\textit{P}$>$0.05)}. This slight lead by Lingshu can be attributed to the disparity in model and data scale, with Lingshu utilizing a larger 7B-parameter model and a more extensive pretraining corpus of approximately 12M samples, compared to our 2B model trained on 0.1M samples. Given that long-contextual reasoning often benefits from larger model, these results suggest that MM-Retinal-Reason and our training method can bridge the capacity gap. Furthermore, since Lingshu's pretraining includes PubMed-based sources, there may also be potential data overlap with our PMC-derived Complex Reasoning test split. 

\subsubsection{Fundamental and Generalizable Representation}
Although FundusExpert achieves competitive performance on CFP basic reasoning tasks (see Tab.~\ref{tab:CFP&FFA}), it underperforms markedly on FFA, OCT basic reasoning tasks, and complex reasoning tasks (see Tab.~\ref{tab:CFP&FFA} \&~\ref{tab:OCT&complex}, and Fig.~\ref{fig:experiment}). This highlights that OphthaReason learned a more fundamental and robust representation of ophthalmic knowledge, indicating advanced ability to reason and diagnose from complex and multi-source clinical information. Moreover, OphthaReason demonstrates strong performance on the OOD benchmark MedXpertQA-MM, achieving the highest average score of 36.07\% among all comparison models. It surpasses {MedGemma by 6.34\%}, HealthGPT by 8.20\%, and Lingshu by 14.75\%, highlighting its robust generalization capability for complex reasoning and its potential for broader clinical applications.

\subsubsection{Performance-Parameter Trade-off}
OphthaReason consistently exhibits clear parameter efficiency in complex reasoning. Despite being 2–4 times smaller than competing models, it achieves results comparable to or superior to those larger counterparts. However, a notable performance gap exists when comparing our model to the proprietary model, such as GPT-4o. We attribute this disparity to the massive scale ($> 100$B\cite{abacha2024medec}) and diversity of the training data used by these models. This extensive and varied exposure is critical for complex reasoning. Nevertheless, deploying such large-scale models in real-world clinical use remains challenging, as most institutions are constrained by the extensive resources required. In contrast, our smaller-scale model offers a more feasible solution for practical adoption.

\subsection{Human and Random Baselines}
To provide interpretable references, we include a random-choice baseline and an ophthalmologist baseline. Given the large scale of the benchmark, {ophthalmologists complete a representative subset of 130 samples constructed via stratified proportional random sampling from MM-Retinal-Reason test set.} For the basic reasoning task, ophthalmologists achieve accuracies of 92.50\%, 94.29\%, and 93.75\% on CFP, FFA, and OCT, respectively, while the random-choice baseline obtains 23.97\%, 27.97\%, and 29.15\%. {In comparison, OphthaReason achieves 70.59\%, 69.50\%, and 65.39\%}, substantially surpassing the random baseline and narrowing the gap with ophthalmologists, indicating that the model has acquired meaningful ophthalmic knowledge beyond statistical guessing. For the complex reasoning task, ophthalmologists achieve an average score of 25.46\%, while OphthaReason reaches 28.31\%, suggesting that this task is challenging even for ophthalmologists and OphthaReason remains effective in multi-image, long-context reasoning involving complex or rare conditions. {The lower panel of Fig.~\ref{fig:case_study} shows a representative case where OphthaReason matches the source-derived reference answer while the ophthalmologist does not, highlighting its ability to integrate ocular findings with systemic and treatment context in challenging differential diagnoses.}

\subsection{Performance on Real-World Clinical Reasoning Tasks}
To further evaluate the clinical utility, we conduct an independent assessment on an in-house clinical reasoning VQA dataset collected from the hospital. {This cohort preserves the heterogeneity of clinical settings to mirror realistic deployment, and OphthaReason is frozen before evaluation to perform blinded inference on real patient cases.} 
This benchmark assesses multimodal reasoning capabilities by requiring the model to integrate diverse ophthalmic imaging modalities with patient clinical information. Specifically, it encompasses 16 ophthalmic conditions spanning retinal, uveal, optic nerve, lens, degenerative, systemic disease–related, and congenital disorders, such as age-related macular degeneration, diabetic retinopathy, retinal detachment, Vogt–Koyanagi–Harada disease, glaucoma, and cataract. As shown in Fig.~\ref{fig:clinical_reasoning}(a), OphthaReason-Intern achieves 45.45\%, outperforming competitive baselines, including medical generalists and ophthalmic specialists. The results suggest that OphthaReason learns transferable diagnostic knowledge that generalizes from public training sources to real-world clinical practice. These can be attributed to multi-source heterogeneous training, which exposes model to diverse clinical scenarios and improves robustness against distribution shifts. Moreover, clinician-authored PMC-derived complex 
reasoning VQA provides authentic patient cases and expert diagnostic logic for training.

\subsection{Evaluation of Reasoning Processes}
\subsubsection{Quantitative Expert Evaluation}
Three ophthalmologists evaluate {120 cases} per model for OphthaReason and Lingshu, {selected from the MM-Retinal-Reason test split using stratified proportional random sampling} and assessed according to the rubric in Sec.~\ref{sec:CoT_data}. As shown in Fig.~\ref{fig:clinical_reasoning}(b), OphthaReason consistently outperforms Lingshu on basic reasoning VQA. On complex reasoning VQA, it achieves higher correctness, completeness, and safety, while slightly lagging in clinical alignment, primarily due to the capacity limits of smaller models in long-context complex reasoning. Nevertheless, high-quality ophthalmic data and specialized training strategy mitigate this limitation, enabling performance comparable to the larger-scale Lingshu. OphthaReason also shows strong and balanced performance on real-world clinical VQA, indicating its robustness and practical reliability in real clinical scenarios. 

\subsubsection{Qualitative Case Analysis}
{OphthaReason generates detailed and interpretable step-by-step reasoning traces, with representative examples in Fig.~\ref{fig:case_study}. OphthaReason excels at identifying critical visual features. In the basic reasoning case, OphthaReason correctly interprets image content (e.g., \textit{``macular changes, drusen, pigmentary alterations"}), while Lingshu omits key visual evidence (e.g., \textit{Drusen and pigmentary changes are not described}) and treats incorrect features (e.g., \textit{``retinal atrophy and 'bone spicule' appearance"}) as defining characteristics. Moreover, OphthaReason exhibits proficiency in differential diagnosis. In the complex reasoning case, it systematically excludes unlikely conditions by integrating multi-source, clinically relevant evidence (e.g., \textit{``the absence of ... makes ... unlikely}) rather than relying on superficial or insufficient cues to rule out non-specific alternative diagnoses. Overall, these underscore OphthaReason’s ability to generate explainable and clinically relevant diagnostic rationales, ensuring its readiness for real-world decision support.}

\subsubsection{Explanation Faithfulness Analysis}
{We perform faithfulness validation via mistake injection by inserting incorrect statements into intermediate reasoning steps and measuring whether the final diagnosis changes from the original complete CoT answer. The results show that 69.89\% of cases change the final diagnosis, suggesting that the prediction is sensitive to the reasoning and providing empirical support for the faithfulness and interpretability of OphthaReason’s reasoning traces.}

\begin{table}[t]
\centering
\caption{\textbf{Ablation of different training stages and UADT on the MM-Retinal-Reason dataset (\%).} The best results are highlighted in \textbf{bold}.}
\label{tab:ablation}
\footnotesize
\renewcommand{\arraystretch}{1.08}
\setlength{\tabcolsep}{3.0pt}

\resizebox{\linewidth}{!}{
\begin{tabular}{llccccc}
\toprule
\textbf{Base Model} & \textbf{Setting} & \textbf{CFP} & \textbf{FFA} & \textbf{OCT} & \textbf{Complex} & \textbf{AVG} \\
\midrule

\multirow{6}{*}{Qwen2.5-VL-3B} 
& Base model & 30.48 & 29.38 & 34.54 & 11.16 & 26.39 \\
& + SFT & 37.31 & 20.00 & 32.34 & 3.25 & 23.23 \\
& + Stage 3 {(Vanilla GRPO)} & 55.81 & 43.75 & 43.34 & 17.10 & 40.00 \\
& + Stage 2\&3 {(Vanilla GRPO)} & 59.79 & 52.38 & 48.25 & \textbf{23.66} & 46.02 \\
& {+ Stage 1\&2\&3 (Vanilla GRPO)} & 62.07 & \textbf{55.13} & 50.86 & 22.52 & 47.65 \\
& \cellcolor[HTML]{E0FFFF}{+ Stage 1\&2\&3 (GRPO + UADT)} & \cellcolor[HTML]{E0FFFF}\textbf{62.93} & \cellcolor[HTML]{E0FFFF}54.88 & \cellcolor[HTML]{E0FFFF}\textbf{55.35} & \cellcolor[HTML]{E0FFFF}23.52 & \cellcolor[HTML]{E0FFFF}\textbf{49.17} \\
\midrule

\multirow{6}{*}{{InternVL3-2B}}
& Base model & 45.23 & 35.25 & 31.37 & 15.53 & 31.85 \\
& + SFT & 55.52 & 46.13 & 45.52 & 14.94 & 40.53 \\
& + Stage 3 {(Vanilla GRPO)} & 63.94 & 55.00 & 58.64 & 17.69 & 48.82 \\
& + Stage 2\&3 {(Vanilla GRPO)} & 66.56 & \textbf{68.15} & 65.68 & 23.34 & 55.93 \\
& {+ Stage 1\&2\&3 (Vanilla GRPO)} & 67.81 & 67.35 & 67.59 & 22.88 & 56.41 \\
& \cellcolor[HTML]{E0FFFF}{+ Stage 1\&2\&3 (GRPO + UADT)} & \cellcolor[HTML]{E0FFFF}\textbf{69.43} & \cellcolor[HTML]{E0FFFF}68.00 & \cellcolor[HTML]{E0FFFF}\textbf{69.50} & \cellcolor[HTML]{E0FFFF}\textbf{25.35} & \cellcolor[HTML]{E0FFFF}\textbf{58.07} \\
\bottomrule
\end{tabular}
}
\end{table}

\begin{figure}[]
\centerline{\includegraphics[scale=0.28]{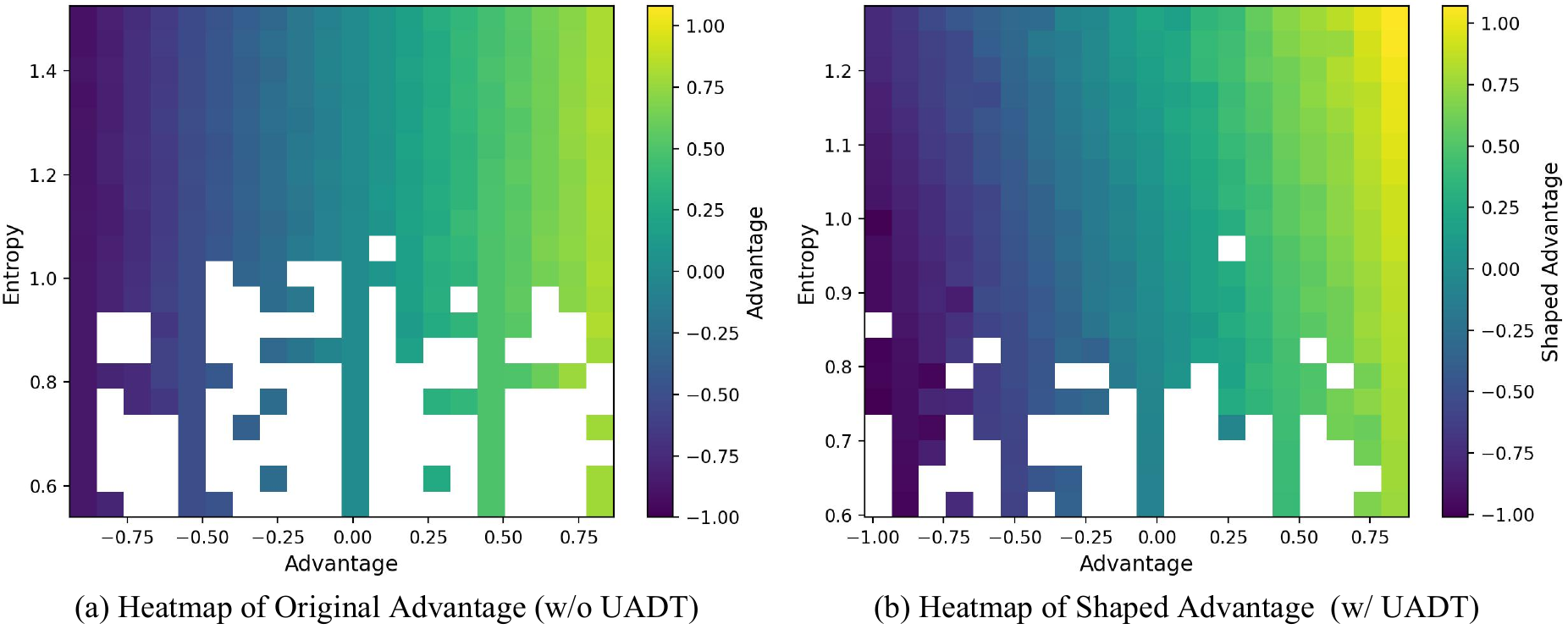}}
\caption{\textbf{Heatmaps of advantage with and without UADT.} (a) Distribution of original advantage across entropy without UADT. (b) Distribution of shaped advantage after applying UADT. The color indicates the magnitude of the advantage, and white areas represent no data.}
\label{fig:ablation}
\end{figure}

\subsection{Ablation Study}
To validate the effectiveness of the three-stage training and the Uncertainty-Aware Dynamic Thinking (UADT), we conduct ablation studies by incrementally adding each component to the base model. The results are summarized in Tab.~\ref{tab:ablation}.

\subsubsection{Impact of Different Training Stages}
For Qwen2.5-VL-3B, naive SFT that supervises final answers reduces the average score from 26.39\% to 23.23\%, suggesting that it may encourage pattern memorization rather than explicit reasoning. In contrast, progressively applying the training stages improves model performance. Stage 3 {with vanilla GRPO} introduces a major training paradigm shift to reinforcement learning and yields the largest gain of 13.61\% over the base model, while Stage 2 activates the model's reasoning capability through CoT supervision and raises the performance by 6.02\%. Notably, UADT further increases performance from 47.65\% to 49.17\% over the established three-stage baseline through adaptive exploration regulation. This gain is achieved with negligible additional computational overhead, efficiently maximizing the model’s reasoning potential without requiring extra data or new paradigms. {To verify consistency across backbones, we further conduct ablation on InternVL3-2B, where the full three-stage pipeline improves the average score from 31.85\% to 56.41\%, and UADT further increases it to 58.07\%. Although UADT shows a more pronounced gain on OCT for Qwen2.5-VL-3B, the gains are more balanced on InternVL3-2B, suggesting that this is mainly due to backbone initial capabilities and modality-specific improvement space rather than a method preference. Overall, these results demonstrate consistent improvements across backbones and modalities.}

\begin{table}[t]
\centering
\caption{{Clinical reader study results under Human-alone, AI-alone, and Human-AI Copilot settings.}}
\label{tab:reader_study}
\renewcommand{\arraystretch}{1.1}
\setlength{\tabcolsep}{4.5pt}
\begin{tabular}{cccc}
\toprule
\textbf{Metric} & \textbf{Human-alone} & \textbf{AI-alone} & \textbf{Human-AI Copilot} \\
\midrule
Diagnostic Accuracy & 56.00\% & 59.33\% & 64.67\% \\
Reasoning Correctness & 61.00\% & 67.50\% & 73.79\% \\
\bottomrule
\end{tabular}
\end{table}

\begin{figure}[]
\centerline{\includegraphics[scale=0.28]{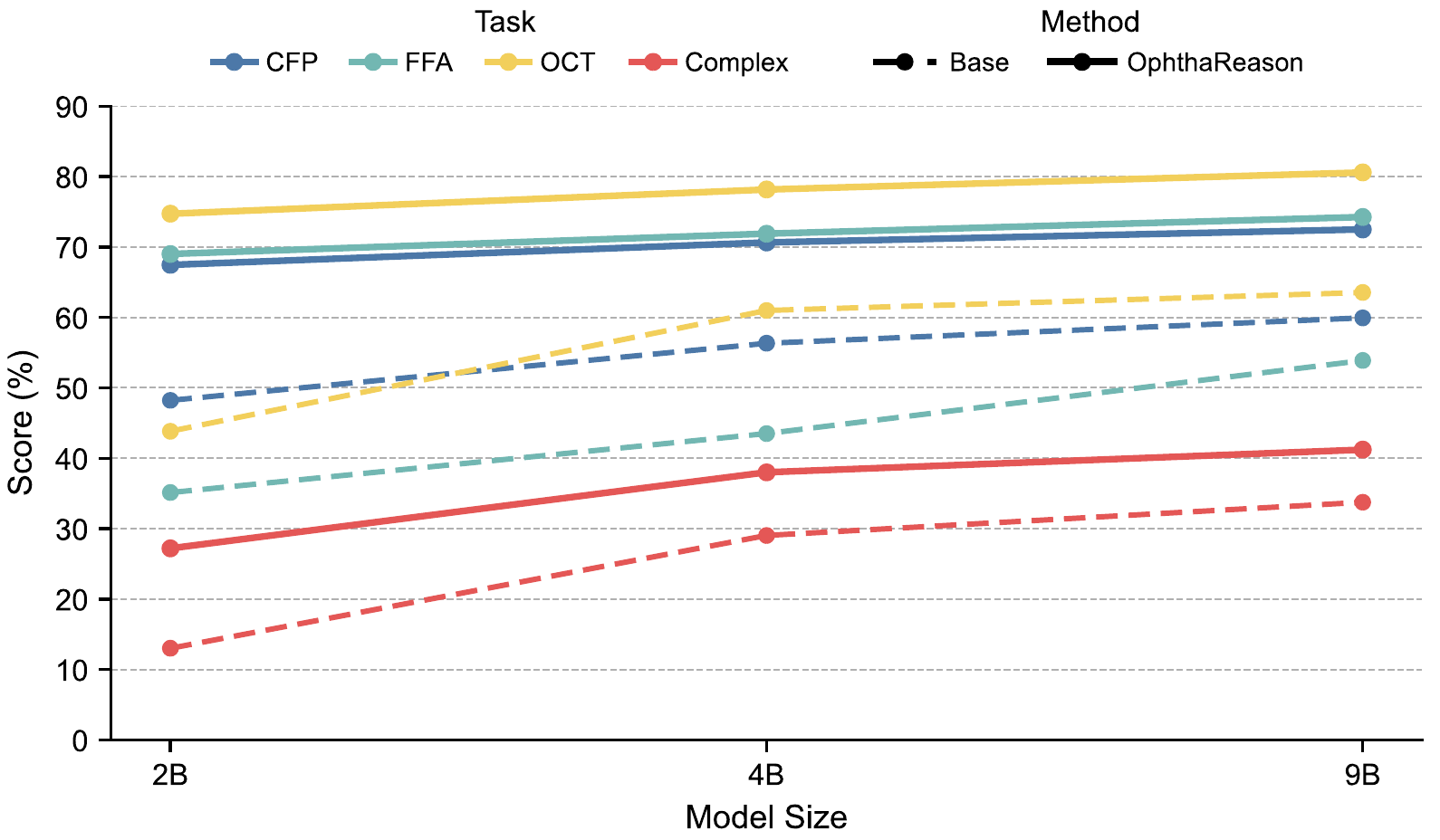}}
\caption{\textbf{Scaling analysis on Qwen3.5 backbones.} Dashed lines denote vanilla Qwen3.5 models, and solid lines denote OphthaReason models.}
\label{fig:qwen35_scaling}
\end{figure}

\subsubsection{Impact of Uncertainty-Aware Dynamic Thinking}
As shown in the last two rows of Tab.~\ref{tab:ablation}, UADT consistently improves performance across tasks, demonstrating its criticality in model performance. To further investigate its mechanism, we visualize the advantage heatmap in Fig.~\ref{fig:ablation}. As shown in the right panel, UADT dynamically modulates advantages based on sample entropy. For high-entropy samples, UADT amplifies the advantage to encourage exploration, evidenced by the color shift from blue/green to brighter green/yellow. Conversely, for low-entropy samples, UADT reduces advantages, causing the corresponding regions to darken. Moreover, the comparison of the two panels reveals that UADT increases sample density in the low-entropy region, particularly in the high-advantage area therein. This redistribution demonstrates that UADT guides the model to produce correct answers with higher confidence, improving reliability and factuality.

\subsubsection{Discussion of Data and Optimization Contributions}
{OphthaReason benefits from the indispensable and complementary roles of data and optimization. MM-Retinal-Reason provides clinically meaningful ophthalmic knowledge and task supervision. Vision-language alignment strengthens ophthalmic visual understanding for subsequent reasoning, while reasoning supervision promotes structured evidence integration and differential reasoning. GRPO enhances the reasoning policy through reward-guided exploration, and UADT further adapts policy updates according to sample uncertainty.}

\subsection{Scalability Analysis}
{To evaluate the scalability of OphthaReason on stronger contemporary backbones, we perform scaling experiments on the Qwen3.5 series, including 2B, 4B, and 9B models. As shown in Fig.~\ref{fig:qwen35_scaling}, OphthaReason consistently improves the vanilla Qwen3.5 models across all model scales, demonstrating that our method is not confined to small or specific backbones, but remains robust on stronger contemporary models. Moreover, OphthaReason's performance improves as the backbone size increases, indicating a positive scaling trend. These results indicate that the three-stage training pipeline and UADT mechanism continue to provide benefits when the base model becomes larger and stronger, supporting the scalability and future-readiness of OphthaReason.}

\subsection{Clinical Reader Study and AI-Copilot Validation}

\begin{figure}[]
\centerline{\includegraphics[scale=0.56]{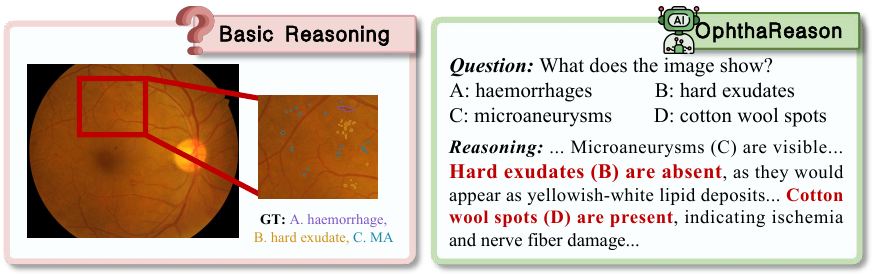}}
\caption{\textbf{Hallucination analysis.} Colored annotations mark the target lesions, and red text highlights the hallucinated model output.}
\label{fig:hallucination}
\end{figure}

{To reflect realistic clinical workflows and assess clinical utility in Computer-Aided Diagnosis, we conduct a clinician-in-the-loop reader study on {150 unseen challenging cases from MM-Retinal-Reason test split selected under expert guidance}. The study includes three cohorts: (1) Human-alone, where two junior clinicians reason and diagnose independently; (2) AI-alone, where OphthaReason independently generates diagnoses and reasoning; and (3) Human-AI Copilot, where two junior clinicians complete diagnoses and reasoning with OphthaReason assistance. Senior ophthalmic experts evaluate diagnostic accuracy by comparing diagnoses against expert references and assess clinical correctness of reasoning with reference to AAO guidelines and Chinese Medical Association Ophthalmology guidelines. The results are shown in Tab.~\ref{tab:reader_study}. In Human-vs-AI comparison, OphthaReason outperforms clinicians in diagnostic accuracy and reasoning correctness, indicating higher standalone model performance. In Human+AI setting, AI-Copilot further improves diagnosis and reasoning correctness, showing that OphthaReason provides additional benefit as a clinical copilot. These results demonstrate the clinical utility of OphthaReason in realistic workflows, both as a standalone diagnostic model and as an AI copilot that improves diagnostic accuracy and reasoning quality.}

\subsection{Limitations and Future Work}
\subsubsection{Failure and Hallucination Cases Analysis}

To provide a transparent evaluation, we analyze representative failures involving incomplete reasoning despite correct answers and {hallucinations causing incorrect predictions}. Even when the diagnosis is correct, the rationale may rely on shortcuts, such as emphasizing coarse anatomical markers like ``enlarged cup-to-disc ratio'' instead of explicitly integrating all evidence, or reaching the answer by eliminating alternatives without describing subtle findings such as tiny hemorrhages. {More critically, the model may contradict visual or textual evidence and propagate the error to its final prediction. As shown in Fig.~\ref{fig:hallucination}, the model fails to identify the annotated hard exudates while hallucinating cotton wool spots that are not supported by the image or reference annotations. This combination of lesion omission and fabrication may underestimate diabetic retinopathy severity, overstate retinal ischemia, and lead to inappropriate referral urgency or follow-up decisions.}

\subsubsection{Limitations and Future Work}

OphthaReason focuses on single-timepoint reasoning without explicitly modeling longitudinal information across visits, which is important for assessing disease progression and treatment response. {The evaluation framework also has limitations. For open-ended responses, lexical overlap metrics are scalable but sensitive to wording and do not directly reflect reasoning validity, while clinician evaluation better captures clinical correctness and safety but remains retrospective and sample-based. Moreover, current perturbation tests cannot fully establish the causal faithfulness of reasoning traces. This remains an open challenge for LLMs and MLLMs and is particularly important in ophthalmology, where unfaithful rationales may incorrectly attribute predictions to ocular findings or clinical information even when they are driven by spurious patterns or irrelevant cues, thereby obscuring decision basis and reducing auditability.}

Future work will incorporate temporal modeling, larger-scale longitudinal data, and prospective real-world validations. {Further studies are also needed to investigate reasoning faithfulness through counterfactual interventions, reasoning perturbations, and mechanistic analyses, develop faithfulness-aware training, and assess how faithful and unfaithful reasoning affect diagnostic decisions and clinician reliance through reader studies.} Broader medical-domain extension and finer-grained visual reasoning also warrant further investigation.

\section{Conclusion}
We introduce MM-Retinal-Reason, a multimodal ophthalmic dataset spanning from visual perception to cognitive reasoning. We also propose OphthaReason, an RL-enhanced ophthalmic multimodal reasoning model trained via a three-stage pipeline. To accommodate varying task complexities and simulate progressive reasoning essential in clinical diagnostics, we design an Uncertainty-Aware Dynamic Thinking (UADT) mechanism. OphthaReason outperforms leading generalist and medical specialist models and produces high-quality reasoning across imaging modalities and clinical scenarios, demonstrating its effectiveness for ophthalmic decision support. Beyond ophthalmology, our data construction pipeline and training framework are transferable to other medical specialties, highlighting the generalizability of our approach.

\bibliographystyle{IEEEtran}
\bibliography{tmi.bib}

@article{jiang2025hulu,
  title={Hulu-med: A transparent generalist model towards holistic medical vision-language understanding},
  author={Jiang, Songtao and Wang, Yuan and Song, Sibo and Hu, Tianxiang and Zhou, Chenyi and Pu, Bin and Zhang, Yan and Yang, Zhibo and Feng, Yang and Zhou, Joey Tianyi and others},
  journal={arXiv preprint arXiv:2510.08668},
  year={2025}
}

@article{sellergren2025medgemma,
  title={Medgemma technical report},
  author={Sellergren, Andrew and Kazemzadeh, Sahar and Jaroensri, Tiam and Kiraly, Atilla and Traverse, Madeleine and Kohlberger, Timo and Xu, Shawn and Jamil, Fayaz and Hughes, C{\'\i}an and Lau, Charles and others},
  journal={arXiv preprint arXiv:2507.05201},
  year={2025}
}

@article{huang2025vision,
  title={Vision-r1: Incentivizing reasoning capability in multimodal large language models},
  author={Huang, Wenxuan and others},
  journal={arXiv preprint arXiv:2503.06749},
  year={2025}
}

@inproceedings{tan2025reason,
  title={Reason-rft: Reinforcement fine-tuning for visual reasoning of vision language models},
  author={Tan, Huajie and Ji, Yuheng and Hao, Xiaoshuai and Chen, Xiansheng and Wang, Pengwei and Wang, Zhongyuan and Zhang, Shanghang},
  booktitle={The Thirty-ninth Annual Conference on Neural Information Processing Systems},
  year={2025}
}

@article{deng2025openvlthinker,
  title={Openvlthinker: An early exploration to complex vision-language reasoning via iterative self-improvement},
  author={Deng, Yihe and Bansal, Hritik and Yin, Fan and Peng, Nanyun and Wang, Wei and Chang, Kai-Wei},
  journal={arXiv preprint arXiv:2503.17352},
  year={2025}
}

@article{guo2025deepseek,
  title={Deepseek-r1: Incentivizing reasoning capability in llms via reinforcement learning},
  author={Guo, Daya and others},
  journal={arXiv preprint arXiv:2501.12948},
  year={2025}
}

@article{jaech2024openai,
  title={Openai o1 system card},
  author={Jaech, Aaron and others},
  journal={arXiv preprint arXiv:2412.16720},
  year={2024}
}

@article{liu2025visual,
  title={Visual-rft: Visual reinforcement fine-tuning},
  author={Liu, Ziyu and others},
  journal={Proceedings of the IEEE/CVF international conference on computer vision},
  year={2025}
}

@article{lai2025med,
  title={Med-r1: Reinforcement learning for generalizable medical reasoning in vision-language models},
  author={Lai, Yuxiang and Zhong, Jike and Li, Ming and Zhao, Shitian and Yang, Xiaofeng},
  journal={arXiv preprint arXiv:2503.13939},
  year={2025}
}

@article{fan2025chestx,
  title={ChestX-Reasoner: Advancing Radiology Foundation Models with Reasoning through Step-by-Step Verification},
  author={Fan, Ziqing and Liang, Cheng and Wu, Chaoyi and Zhang, Ya and Wang, Yanfeng and Xie, Weidi},
  journal={arXiv preprint arXiv:2504.20930},
  year={2025}
}

@article{zhang2025patho,
  title={Patho-R1: A Multimodal Reinforcement Learning-Based Pathology Expert Reasoner},
  author={Zhang, Wenchuan and others},
  journal={arXiv preprint arXiv:2505.11404},
  year={2025}
}

@article{thapa2025disentangling,
  title={Disentangling reasoning and knowledge in medical large language models},
  author={Thapa, Rahul and others},
  journal={arXiv preprint arXiv:2505.11462},
  year={2025}
}

@article{pfob2022importance,
  title={The importance of multi-modal imaging and clinical information for humans and AI-based algorithms to classify breast masses (INSPiRED 003): an international, multicenter analysis},
  author={Pfob, Andr{\'e} and others},
  journal={European radiology},
  volume={32},
  number={6},
  pages={4101--4115},
  year={2022},
  publisher={Springer}
}

@inproceedings{ben2019vqa,
  title={Vqa-med: Overview of the medical visual question answering task at imageclef 2019},
  author={Ben Abacha, Asma and Hasan, Sadid A and Datla, Vivek V and Demner-Fushman, Dina and M{\"u}ller, Henning},
  booktitle={Proceedings of CLEF (Conference and Labs of the Evaluation Forum) 2019 Working Notes},
  year={2019}
}

@article{zhang2023pmc,
  title={Development of a large-scale medical visual question-answering dataset},
  author={Zhang, Xiaoman and others},
  journal={Communications Medicine},
  volume={4},
  number={1},
  pages={277},
  year={2024},
  publisher={Nature Publishing Group UK London}
}

@inproceedings{hu2024omnimedvqa,
  title={Omnimedvqa: A new large-scale comprehensive evaluation benchmark for medical lvlm},
  author={Hu, Yutao and others},
  booktitle={Proceedings of the IEEE/CVF Conference on Computer Vision and Pattern Recognition},
  pages={22170--22183},
  year={2024}
}

@inproceedings{zuo2025medxpertqa,
  title={MedXpertQA: Benchmarking Expert-Level Medical Reasoning and Understanding},
  author={Zuo, Yuxin and others},
  booktitle={Forty-second International Conference on Machine Learning},
  year={2025}
}

@article{yu2025medframeqa,
  title={MedFrameQA: A Multi-Image Medical VQA Benchmark for Clinical Reasoning},
  author={Yu, Suhao and Wang, Haojin and Wu, Juncheng and Xie, Cihang and Zhou, Yuyin},
  journal={arXiv preprint arXiv:2505.16964},
  year={2025}
}

@article{chao2019bilateral,
  title={Bilateral cytomegalovirus retinitis comorbid with diabetic macular edema},
  author={Chao, Yu-Jang and Huang, Yi-Ming and Chung, Yu-Chien and Hwang, De-Kuang and Chen, Shih-Jen and Liu, Catherine Jui-Ling},
  journal={Taiwan journal of ophthalmology},
  volume={9},
  number={2},
  pages={122--126},
  year={2019},
  publisher={Medknow}
}

@inproceedings{li2025eyecaregpt,
  title={Eyecaregpt: Boosting comprehensive ophthalmology understanding with tailored dataset, benchmark and model},
  author={Li, Sijing and Lin, Tianwei and Lin, Lingshuai and Zhang, Wenqiao and Liu, Jiang and Yang, Xiaoda and Li, Juncheng and He, Yucheng and Song, Xiaohui and Xiao, Jun and others},
  booktitle={Proceedings of the 33rd ACM International Conference on Multimedia},
  pages={3893--3902},
  year={2025}
}

@article{wu2025medcasereasoning,
  title={MedCaseReasoning: Evaluating and learning diagnostic reasoning from clinical case reports},
  author={Wu, Kevin and others},
  journal={arXiv preprint arXiv:2505.11733},
  year={2025}
}

@inproceedings{qin2025lmod,
  title={Lmod: A large multimodal ophthalmology dataset and benchmark for large vision-language models},
  author={Qin, Zhenyue and Yin, Yu and Campbell, Dylan and Wu, Xuansheng and Zou, Ke and Liu, Ninghao and Tham, Yih Chung and Zhang, Xiuzhen Jenny and Chen, Qingyu},
  booktitle={Findings of the Association for Computational Linguistics: NAACL 2025},
  pages={2501--2522},
  year={2025}
}

@inproceedings{chen2024huatuogpt,
  title={Towards medical complex reasoning with LLMs through medical verifiable problems},
  author={Chen, Junying and Cai, Zhenyang and Ji, Ke and Wang, Xidong and Liu, Wanlong and Wang, Rongsheng and Wang, Benyou},
  booktitle={Findings of the Association for Computational Linguistics: ACL 2025},
  pages={14552--14573},
  year={2025}
}

@article{xu2025lingshu,
  title={Lingshu: A Generalist Foundation Model for Unified Multimodal Medical Understanding and Reasoning},
  author={Xu, Weiwen and others},
  journal={arXiv preprint arXiv:2506.07044},
  year={2025}
}

@article{dai2025qoq,
  title={QoQ-Med: Building Multimodal Clinical Foundation Models with Domain-Aware GRPO Training},
  author={Dai, Wei and Chen, Peilin and Ekbote, Chanakya and Liang, Paul Pu},
  journal={arXiv preprint arXiv:2506.00711},
  year={2025}
}

@article{jing2025reason,
    title = {Reason like a radiologist: Chain-of-thought and reinforcement learning for verifiable report generation},
    journal = {Medical Image Analysis},
    volume = {109},
    pages = {103910},
    year = {2026},
    issn = {1361-8415},
    author = {Peiyuan Jing and Kinhei Lee and Zhenxuan Zhang and Huichi Zhou and Zhengqing Yuan and Zhifan Gao and Lei Zhu and Giorgos Papanastasiou and Yingying Fang and Guang Yang},
}

@article{wu2025pathvlm,
  title={PathVLM-R1: A Reinforcement Learning-Driven Reasoning Model for Pathology Visual-Language Tasks},
  author={Wu, Jianyu and others},
  journal={arXiv preprint arXiv:2504.09258},
  year={2025}
}

@inproceedings{chen2024huatuovision,
  title={Towards injecting medical visual knowledge into multimodal llms at scale},
  author={Chen, Junying and others},
  booktitle={Proceedings of the 2024 conference on empirical methods in natural language processing},
  pages={7346--7370},
  year={2024}
}

@inproceedings{huang2025towards,
  title={Towards a multimodal large language model with pixel-level insight for biomedicine},
  author={Huang, Xiaoshuang and others},
  booktitle={Proceedings of the AAAI Conference on Artificial Intelligence},
  volume={39},
  number={4},
  pages={3779--3787},
  year={2025}
}

@inproceedings{li2019attention,
  title={Attention based glaucoma detection: A large-scale database and CNN model},
  author={Li, Liu and Xu, Mai and Wang, Xiaofei and Jiang, Lai and Liu, Hanruo},
  booktitle={Proceedings of the IEEE/CVF conference on computer vision and pattern recognition},
  pages={10571--10580},
  year={2019}
}

@article{pachade2021retinal,
  title={Retinal fundus multi-disease image dataset (rfmid): A dataset for multi-disease detection research},
  author={Pachade, Samiksha and others},
  journal={Data},
  volume={6},
  number={2},
  pages={14},
  year={2021},
  publisher={MDPI}
}

@article{hoover2000locating,
  title={Locating blood vessels in retinal images by piecewise threshold probing of a matched filter response},
  author={Hoover, AD},
  journal={IEEE Transactions on Medical imaging},
  volume={19},
  number={3},
  pages={203--210},
  year={2000},
  publisher={IEEE}
}

@article{hoover2003locating,
  title={Locating the optic nerve in a retinal image using the fuzzy convergence of the blood vessels},
  author={Hoover, Adam and Goldbaum, Michael},
  journal={IEEE transactions on medical imaging},
  volume={22},
  number={8},
  pages={951--958},
  year={2003},
  publisher={IEEE}
}

@inproceedings{
  liu2022dab,
  title={{DAB}-{DETR}: Dynamic Anchor Boxes are Better Queries for {DETR}},
  author={Shilong Liu and others},
  booktitle={International Conference on Learning Representations},
  year={2022},
  url={https://openreview.net/forum?id=oMI9PjOb9Jl}
}

@article{bai2025qwen2,
  title={Qwen2. 5-vl technical report},
  author={Bai, Shuai and others},
  journal={arXiv preprint arXiv:2502.13923},
  year={2025}
}

@inproceedings{wu2024mm,
  title={MM-retinal: Knowledge-enhanced foundational pretraining with fundus image-text expertise},
  author={Wu, Ruiqi and Zhang, Chenran and Zhang, Jianle and Zhou, Yi and Zhou, Tao and Fu, Huazhu},
  booktitle={International Conference on Medical Image Computing and Computer-Assisted Intervention},
  pages={722--732},
  year={2024},
  organization={Springer}
}

@article{wang2025perception,
  title={Perception-Aware Policy Optimization for Multimodal Reasoning},
  author={Wang, Zhenhailong and others},
  journal={arXiv preprint arXiv:2507.06448},
  year={2025}
}

@article{su2025gmai,
  title={Gmai-vl-r1: Harnessing reinforcement learning for multimodal medical reasoning},
  author={Su, Yanzhou and others},
  journal={arXiv preprint arXiv:2504.01886},
  year={2025}
}

@article{wu2025mm,
  title={MM-Retinal V2: Transfer an Elite Knowledge Spark into Fundus Vision-Language Pretraining},
  author={Wu, Ruiqi and others},
  journal={arXiv preprint arXiv:2501.15798},
  year={2025}
}

@article{rui2025improving,
  title={Improving Medical Reasoning with Curriculum-Aware Reinforcement Learning},
  author={Rui, Shaohao and Chen, Kaitao and Ma, Weijie and Wang, Xiaosong},
  journal={arXiv preprint arXiv:2505.19213},
  year={2025}
}

@article{shao2024deepseekmath,
  title={Deepseekmath: Pushing the limits of mathematical reasoning in open language models},
  author={Shao, Zhihong and others},
  journal={arXiv preprint arXiv:2402.03300},
  year={2024}
}

@article{kadavath2022language,
  title={Language models (mostly) know what they know},
  author={Kadavath, Saurav and others},
  journal={arXiv preprint arXiv:2207.05221},
  year={2022}
}

@article{cheng2025reasoning,
  title={Reasoning with exploration: An entropy perspective},
  author={Cheng, Daixuan and others},
  journal={arXiv preprint arXiv:2506.14758},
  year={2025}
}

@inproceedings{pan2025medvlm,
  title={Medvlm-r1: Incentivizing medical reasoning capability of vision-language models (vlms) via reinforcement learning},
  author={Pan, Jiazhen and Liu, Che and Wu, Junde and Liu, Fenglin and Zhu, Jiayuan and Li, Hongwei Bran and Chen, Chen and Ouyang, Cheng and Rueckert, Daniel},
  booktitle={International Conference on Medical Image Computing and Computer-Assisted Intervention},
  pages={337--347},
  year={2025},
  organization={Springer}
}

@inproceedings{kwon2023efficient,
  title={Efficient memory management for large language model serving with pagedattention},
  author={Kwon, Woosuk and others},
  booktitle={Proceedings of the 29th symposium on operating systems principles},
  pages={611--626},
  year={2023}
}

@article{ye2024gmai,
  title={Gmai-mmbench: A comprehensive multimodal evaluation benchmark towards general medical ai},
  author={Ye, Jin and others},
  journal={Advances in Neural Information Processing Systems},
  volume={37},
  pages={94327--94427},
  year={2024}
}

@article{shi2024eyeclip,
  title={A multimodal visual--language foundation model for computational ophthalmology},
  author={Shi, Danli and others},
  journal={npj Digital Medicine},
  volume={8},
  number={1},
  pages={381},
  year={2025},
  publisher={Nature Publishing Group UK London}
}

@article{silva2025foundation,
  title={A foundation language-image model of the retina (flair): Encoding expert knowledge in text supervision},
  author={Silva-Rodriguez, Julio and Chakor, Hadi and Kobbi, Riadh and Dolz, Jose and Ayed, Ismail Ben},
  journal={Medical Image Analysis},
  volume={99},
  pages={103357},
  year={2025},
  publisher={Elsevier}
}

@article{wang2024common,
  title={Enhancing diagnostic accuracy in rare and common fundus diseases with a knowledge-rich vision-language model},
  author={Wang, Meng and others},
  journal={Nature Communications},
  volume={16},
  number={1},
  pages={5528},
  year={2025},
  publisher={Nature Publishing Group UK London}
}

@article{liu2025constructing,
  title={Constructing Ophthalmic MLLM for Positioning-diagnosis Collaboration Through Clinical Cognitive Chain Reasoning},
  author={Liu, Xinyao and Song, Diping},
  journal={Proceedings of the IEEE/CVF International Conference on Computer Vision},
  year={2025}
}

@inproceedings{abacha2024medec,
    title = "{MEDEC}: A Benchmark for Medical Error Detection and Correction in Clinical Notes",
    author={Abacha, Asma Ben and others},
    booktitle = "Findings of the Association for Computational Linguistics: ACL 2025",
    year = "2025",
    pages = "22539--22550"
}

@article{luo2025survey,
  title={A Survey of Multimodal Ophthalmic Diagnostics: From Task-Specific Approaches to Foundational Models},
  author={Luo, Xiaoling and others},
  journal={arXiv preprint arXiv:2508.03734},
  year={2025}
}

@misc{OCTC82021,
  year = {2021},
  url = {Kaggle. https://doi.org/10.34740/KAGGLE/DSV/2736749},
  title = {Retinal OCT Image Classification - C8},
  author = {Obuli Sai Naren}
}

@misc{o4mini,
  year = {Apr 2025},
  url = {https://openai.com/index/introducing-o3-and-o4-mini/},
  title = {Introducing o3 and o4-mini},
  author = {OpenAI}
}

@article{achiam2023gpt,
  title={Gpt-4 technical report},
  author={Achiam, Josh and others},
  journal={arXiv preprint arXiv:2303.08774},
  year={2023}
}

@article{gpt4V,
  title={Gpt-4v(ision) technical work and authors},
  year={2023}
}

@article{chu2010concurrent,
  title={Concurrent central retinal artery occlusion and branch retinal vein occlusion in giant cell arteritis},
  author={Chu, Edward R and Chen, Celia S},
  journal={Clinical Ophthalmology},
  pages={565--567},
  year={2010},
  publisher={Taylor \& Francis}
}

@article{han2021max,
  title={A max-min entropy framework for reinforcement learning},
  author={Han, Seungyul and Sung, Youngchul},
  journal={Advances in Neural Information Processing Systems},
  volume={34},
  pages={25732--25745},
  year={2021}
}

@inproceedings{gaidemystifying,
  title={Demystifying Entropy Control in LLM RL Training: Theoretical Analysis and Dynamic Scheduling},
  author={Gai, Jingchu and others},
  booktitle={Forty-third International Conference on Machine Learning}
}

@inproceedings{haarnoja2018soft,
  title={Soft actor-critic: Off-policy maximum entropy deep reinforcement learning with a stochastic actor},
  author={Haarnoja, Tuomas and Zhou, Aurick and Abbeel, Pieter and Levine, Sergey},
  booktitle={International conference on machine learning},
  pages={1861--1870},
  year={2018},
  organization={Pmlr}
}

@inproceedings{liu2026stable,
  title     = {Stable and Efficient Single-Rollout {RL} for Multimodal Reasoning},
  author    = {Liu, R. and Yu, D. and Ke, L. and others},
  booktitle = {Proceedings of the IEEE/CVF Conference on Computer Vision and Pattern Recognition},
  pages     = {12009--12018},
  year      = {2026}
}

@inproceedings{shen2026entropy,
  title     = {On Entropy Control in {LLM-RL} Algorithms},
  author    = {Shen, Han},
  booktitle = {The Fourteenth International Conference on Learning Representations},
  year      = {2026}
}

@inproceedings{moor2023med,
  title={Med-flamingo: a multimodal medical few-shot learner},
  author={Moor, Michael and Huang, Qian and Wu, Shirley and Yasunaga, Michihiro and Dalmia, Yash and Leskovec, Jure and Zakka, Cyril and Reis, Eduardo Pontes and Rajpurkar, Pranav},
  booktitle={Machine Learning for Health (ML4H)},
  pages={353--367},
  year={2023},
  organization={PMLR}
}

@article{comanici2025gemini,
  title={Gemini 2.5: Pushing the frontier with advanced reasoning, multimodality, long context, and next generation agentic capabilities},
  author={Comanici, Gheorghe and others},
  journal={arXiv preprint arXiv:2507.06261},
  year={2025}
}

@misc{li2024llava,
    title={LLaVA-NeXT: Improved reasoning, OCR, and world knowledge},
    url={https://llava-vl.github.io/blog/2024-01-30-llava-next/},
    author={Liu, Haotian and Li, Chunyuan and Li, Yuheng and Li, Bo and Zhang, Yuanhan and Shen, Sheng and Lee, Yong Jae},
    month={January},
    year={2024}
}

@misc{chen2024internvl,
      title={InternVL3: Exploring Advanced Training and Test-Time Recipes for Open-Source Multimodal Models}, 
      author={Jinguo Zhu and others},
      year={2025},
      eprint={2504.10479},
      archivePrefix={arXiv},
      primaryClass={cs.CV},
      url={https://arxiv.org/abs/2504.10479}, 
}

@article{li2023llava,
  title={Llava-med: Training a large language-and-vision assistant for biomedicine in one day},
  author={Li, Chunyuan and others},
  journal={Advances in Neural Information Processing Systems},
  volume={36},
  pages={28541--28564},
  year={2023}
}

@inproceedings{lin2025healthgpt,
  title={HealthGPT: A Medical Large Vision-Language Model for Unifying Comprehension and Generation via Heterogeneous Knowledge Adaptation},
  author={Lin, Tianwei and others},
  booktitle={Forty-second International Conference on Machine Learning}
}

\end{document}